\documentclass[12pt]{iopart}
\pdfoutput=1
\usepackage{xcolor}
\usepackage{amsfonts}
\usepackage{bm}
\usepackage{graphicx}
\usepackage{subfigure}
\usepackage{algorithm}
\usepackage[noend]{algpseudocode}
\usepackage{comment}

\usepackage[square,compress,numbers]{natbib}
\newcommand{\eqref}[1]{(\ref{#1})}

\begin{document}
\title{Compressed sensing reconstruction using Expectation Propagation}

\author{Alfredo Braunstein$^{1,2,3,4}$, Anna Paola Muntoni$^{1,5,6}$, Andrea Pagnani$^{1,2,4}$ and
Mirko Pieropan$^1$\footnote{Author to whom any correspondence should be addressed}}

\address{$^1$ Department of Applied Science and Technology (DISAT), Politecnico di Torino,
Corso Duca degli Abruzzi 24, Torino, Italy}
\address{$^2$ Italian Institute for Genomic Medicine (IIGM) (former HuGeF),
Via Nizza 52, Torino, Italy}
\address{$^3$ Collegio Carlo Alberto, Via Real Collegio 30, Moncalieri, Italy}
\address{$^4$ INFN Sezione di Torino, Via P. Giuria 1, Torino, Italy}
\address{$^5$ Laboratoire de Physique de l'Ecole Normale Sup\'erieure, ENS, Universit\'e PSL, CNRS, Sorbonne Universit\'e, Universit\'e de Paris, Paris, France}
\address{$^6$ Sorbonne Universit\' e, CNRS, Institut de Biologie Paris-Seine, Laboratory of Computational and Quantitative Biology, F-75005, Paris, France}

\ead{mirko.pieropan@polito.it}

\begin{abstract}

Many interesting problems in fields ranging from telecommunications to
computational biology can be formalized in terms of large
underdetermined systems of linear equations with additional
constraints or regularizers. One of the most studied ones, the Compressed
Sensing problem (CS), consists in finding the solution with the
smallest number of non-zero components of a given system of linear
equations $\bm y = \mathbf{F} {\bm w}$ for known measurement vector
$\bm y$ and sensing matrix $\mathbf{F}$.  Here, we will address the
compressed sensing problem within a Bayesian inference framework where
the sparsity constraint is remapped into a singular prior distribution
(called Spike-and-Slab or Bernoulli-Gauss). Solution to the problem is
attempted through the computation of marginal distributions via
Expectation Propagation (EP), an iterative computational scheme
originally developed in Statistical Physics. We will show that this
strategy is more accurate for statistically correlated measurement matrices.
For computational strategies based on the Bayesian framework
such as variants of Belief Propagation, this is to be expected, as
they implicitly rely on the hypothesis of statistical independence
among the entries of the sensing matrix. Perhaps surprisingly, the
method outperforms uniformly also all the other state-of-the-art methods
in our tests.

\end{abstract}

\maketitle

\section{Introduction}

The problem of Compressed Sensing (CS)
\cite{candes2005,donoho2006,candes2008} has led to significant
developments in the field of sparse approximation and representation
\cite{zhang2015}, together with the more traditional framework of
optimal signal processing \cite{zhang2015,mishali2011}.

In general, CS deals with the reconstruction of a sparse
$N-$dimensional signal from $M$, often noisy, measurements. In this
context, a sparse signal is characterized by a large (possibly the
largest) number of components equal to zero. In practical cases, the
regime of interest is given by the $M\ll N$, $K\ll N$ limit of the
problem, where $K$ is the number of non-zero components. A brute-force
exhaustive search of the $K-$dimensional minimal support of the
$N-$dimensional signal would lead to an exponential explosion in the
search space as there are $N\choose K$ possible base supports to be
explored. In the noiseless limit of CS, even if the $K$-dimensional
minimal base were given, at least $M=K$ measurements are necessary to
uniquely identify the solution.

CS can be easily formulated: let $\bm w \in {\mathbb R}^N$ be
compressed into a vector $\bm y\in {\mathbb R}^M$ (from now on we will
assume $M<N$) through a linear transformation:
\begin{equation}
\bm y = \mathbf{F}  {\bm w},
\label{eq:y=Aw}
\end{equation}
where $\mathbf{F}\in{\mathbb R}^{M\times N}$ is a linear operator of
maximal rank often referred to as the \emph{measurement} or
\emph{sensing} matrix. The problem is to determine the vector $\bm w$
from the knowledge of the measurement matrix $\mathbf{F}$ and of the
compressed vector $\bm y$.

The relevance of CS in statistical physics stems from
the link with the statistical mechanics of disordered systems, in analogy
with many other combinatorial optimization problems
\cite{Mezard_Parisi_Zecchina,mulet2002,mezardbook2009}. In
particular CS is an optimization problem with quenched disorder (the
measurement matrix) and, as such, it is amenable to analytic treatment using
replica theory \cite{kabashima} as developed in the field of
spin-glasses \cite{Parisi_SGB}.

From equation~\eqref{eq:y=Aw}, it is clear that, as long as $M<N$,
there are infinitely many solutions to the system of
equations. However, if one imposes the supplementary condition on the
sparsity of the signal $\bm w$, solving the problem may still lead to
a unique retrieved signal. Thus, a problem in CS is to determine the
set of conditions under which it is possible to find a solution of
equation~\eqref{eq:y=Aw} that is as sparse as possible.  A practical
way to enforce sparsity is through the minimization of the $L_p$-norm
of the $\bm w$ vector in the space of solutions which is defined in
equation~\eqref{eq:y=Aw}. A particularly successful line of research
has been pursued through the minimization of the $L_1$ norm
\cite{candes2005, candes2006, candes2008, candes2008bis,
  donoho2006,donoho2009mp,kabashima}, for which the following
prediction was obtained in the case of independent and identically
distributed (i.i.d.) random Gaussian measurement matrices: in the
large $N$ limit, there is a non-trivial region $\rho = K/N$,
$\alpha=M/N$ where an exact reconstruction of the original signal is
indeed possible. The parameters $\rho$ and $\alpha$ that characterize
the signal are called the \emph{signal density} and the
\emph{measurement rate}, respectively.

However, in many applications, the sensing matrix may not be
random \cite{det_sens_mat} or the i.i.d. assumption might not hold
and, in these cases, several algorithms could fail to decode the
signal.  Examples of deterministic matrices include chirp sensing matrices
\cite{APPLEBAUM2009283}, which have been applied to image
reconstruction \cite{chirp-image}, and second-order Reed-Muller
matrices \cite{4558486}, whereas examples of correlated random
matrices include random partial Fourier matrices
\cite{4016283,1580791}, which are encountered in MRI \cite{mri} as well as in
other applications \cite{magli,radar}, and partial random circulant
and Toeplitz matrices \cite{Rauhut2009CirculantAT}, which arise in the
presence of convolutions.

Our approach here, is to treat CS as a Bayesian inference
problem, i.e. we focus our attention on determining the probability of
observing, \emph{a posteriori}, the signal $\bm{w}$ when we have
observed a set of measurements $\bm{y}$ and given the sensing matrix
$\mathbf{F}$. While equation \eqref{eq:y=Aw} is easily cast in a
Gaussian-like likelihood function, the prior over the variables
$\bm{w}$, enforcing the sparsity constraint, plays a key role on the
goodness of the solution and on the complexity of the problem. A prior
that corresponds to the minimization of a $L_p$ norm, with $p = {1,
  2}$, allows an easy marginalization of the corresponding posterior
probability with the drawback of softening the sparsity
requirement. Similarly to \cite{kabashima,PhysRevX.2.021005,
  krzakala}, we consider the so-called spike-and-slab prior
\cite{spikeandslabprior} which is understood as the minimization of
the $L_0$ norm of the vector $\bm{w}$. The posterior distribution
results to be intractable in practice and thus some approximations
need to be sought.

Almost twenty years ago Expectation Propagation (EP), an iterative
scheme to approximate intractable distributions, has been introduced
first in the field of statistical physics
\cite{opper_gaussian_2000,opper_adaptive_2001} and shortly after in
the field of theoretical computer science
\cite{minka_expectation_2001}. Recently, EP inspired inference
strategies -- similar to the one we present here -- have been proposed to
solve other underdetermined linear constraint problems such as the
problem of sampling solutions from the reconstruction of large scale metabolic
networks \cite{braunstein2017} and of tomographic images
\cite{muntoni2018non}. Here we propose an efficient and accurate
EP-based reconstruction strategy for CS which, moreover, does not require
i.i.d. measurement matrices.

Other attempts to solve sparse linear models using the EP
approximation can be found in \cite{seeger_bayesian_2008} -- where the
authors use a similar EP implementation to the one we adopt here and
make use of a Laplace prior -- and in \cite{Hernandez-Lobato2015} --
where, in addition to the spike-and-slab, a Bernoulli prior on the
components of the signal is introduced and the EP update scheme
involves only three approximating factors of the original posterior
distribution.

All these approaches have the same computational complexity, which is
dominated by a single matrix inversion per EP cycle. In our
implementation, we integrate the inference of the signal with a
variational learning of the density parameter of the spike-and-slab
prior. Besides, we show in \ref{app:T=0} that the original EP scheme
is equivalent to an alternative formulation of the EP update equations
that takes into account the linear constraints \eqref{eq:y=Aw}
exactly. This implementation has the advantage of reducing the size of
the matrix to be inverted, thus decreasing the computing time.  Throughout
the work presented in this paper, we have numerically checked that
both procedures can be used interchangeably in the case of noiseless
measurements, as they lead to the same results.

In conclusion, we believe that our EP algorithm shows some original features
that have not been proposed by other methods, namely: (i) the quality of EP
reconstruction on correlated measurement matrices is the same as in the case of
uncorrelated measurements whereas all other methods we tested fail; (ii) to
reconstruct signals in the noiseless case, we introduced an EP formulation
performing {\it analytically} the zero temperature limit; (iii) we learn the
density parameter of the spike and slab prior from the data by minimizing
iteratively the EP free energy.

The outline is as follows: we introduce the EP approach to CS and set
the basic notations in section \ref{sec:ep}, we show a thorough
comparison with other state of the art algorithms in section
\ref{sec:exp} and, finally, we present our conclusions in section
\ref{sec:concl}.

\section{Expectation Propagation}
\label{sec:ep}

In this section we introduce the so-called finite temperature
formulation of EP, where we allow the measurement vector $\bm{y}$ to
be noisy. We consider the \emph{undersampling} regime $M<N$, although there is no technical limitation in considering
the $M\ge N$ case. The linear constraints in equation \eqref{eq:y=Aw}
can be alternatively mapped into a minimization problem of the
following quadratic form:
\begin{equation}
E(\bm{w})=\frac{1}{2}\left\Vert {\bm y}-\mathbf{F}\boldsymbol{w}\right\Vert _{2}:=\frac{1}{2}\left({\bm y}-\mathbf{F}\boldsymbol{w}\right)^{T}\cdot\left({\bm y}-\mathbf{F}\boldsymbol{w}\right),\label{eq:energy}
\end{equation}
From a Bayesian perspective, the probability of observing the vector
${\bm y}$ given the matrix $\mathbf{F}$ and the vector ${\bm w}$ is:
\begin{eqnarray}
P({\bm y}|\mathbf{F},{\bm w}) & = & \left(\frac{\beta}{2\pi}\right)^{\frac{M}{2}}e^{-\beta E(\bm{w})},\label{eq:py}
\end{eqnarray}
where we introduced a fictitious inverse temperature $\beta$ that we
can take as large as we wish in order to enforce the linear bounds
expressed by equation \eqref{eq:y=Aw}. Alternatively we can interpret
equation \eqref{eq:py} as the probability of observing an additive
noise vector $\bm{y}- \mathbf{F}\bm{w}$ whose elements are distributed
according to a Gaussian density of zero mean and variance
$\beta^{-1}$. The posterior distribution for the vector ${\bm w}$
reads:
\begin{eqnarray}
P\left(\boldsymbol{w}|\mathbf{F},\boldsymbol{y}\right) & = & \frac{P\left(\boldsymbol{y}|\mathbf{F},{\bm w}\right)P\left(\boldsymbol{w}\right)}{P\left({\bm y}\right)}
 =
 \frac{1}{Z_{P}}e^{-\beta\frac{\left({\bm y}-\mathbf{F}\boldsymbol{w}\right)^{T}\cdot\left({\bm y}-\mathbf{F}\boldsymbol{w}\right)}{2}}\prod_{i=1}^{N}\psi_{i}(w_{i}),\label{eq:posterior}
\end{eqnarray}
where in the last step we restricted the structure of the prior
$P({\bm w}):=\prod_{i=1}^{N}\psi_{i}(w_{i})$ to a factorized form,
although more general structures can be considered, e.g. as in
\cite{braunstein2018loop}. In this work, we have considered the
so-called spike-and-slab prior \cite{spikeandslabprior}:
\begin{equation}
\psi(w_i)=(1-\rho)\delta(w_i)+\frac{\rho}{\sqrt{2\pi\lambda}}e^{-\frac{w_i^2}{2\lambda}}, \label{eq:l0}
\end{equation}
where $\delta$ denotes the Dirac delta function, in order to model any prior knowledge about the sparsity of the signal.

We seek a solution vector $\hat{\bm{w}}$ whose components are the
first moments of the marginal densities of equation
\eqref{eq:posterior}. Contrarily to the maximum a posteriori estimate,
it can be proven that this strategy minimizes the mean squared error
between the true and the recovered signal. Unfortunately, due to the
non-convex nature of the spike-and-slab prior, there exists no
technique able to perform the marginalization of the posterior
probability in equation \eqref{eq:posterior}. In the following we
introduce the EP approximation scheme which relies on an adaptive
Gaussian approximation of the marginal probabilities of interest.

\subsection{The approximate posterior distribution}

EP \cite{minka_expectation_2001} is an efficient approximation to
compute posterior probabilities. EP was first introduced as an
improved mean-field method
\cite{opper_gaussian_2000,opper_adaptive_2001} and further developed
in the framework of Bayesian inference problems in the seminal work of
Minka \cite{minka_expectation_2001}. The approximated distribution
consists in substituting the typically analytically intractable
$\psi_{i}$ priors with univariate Gaussian distributions
$\phi_{i}(w_{i})={\cal N}(w_{i};a_{i},d_{i})$ of mean $a_{i}$ and
variance $d_{i}.$ The approximated posterior thus reads:
\begin{eqnarray}
Q({\bm w}|\mathbf{F},{\bm y})& = &\frac{1}{Z_{Q}}e^{-\beta\frac{\left({\bm y}-\mathbf{F}\boldsymbol{w}\right)^{T}\cdot\left({\bm y}-\mathbf{F}\boldsymbol{w}\right)}{2}}\prod_{i=1}^{N}\phi_{i}(w_{i})\nonumber\\
&:=&\frac{1}{Z_{Q}}e^{-\frac{1}{2}({\bm w}-\bar{{\bm w}})^{T}\bm\Sigma^{-1}({\bm w}-\bar{{\bm w}})},\label{eq:approx_Q}
\end{eqnarray}
where:
\begin{equation}
Z_{Q}=\left(2\pi\right)^{\frac{N}{2}}\left(\det\bm\Sigma\right)^{\frac{1}{2}},
\end{equation}
\begin{equation}
\bm\Sigma^{-1}:=\beta \mathbf{F}^{T}\mathbf{F}+\bm D,\quad\bar{{\bm w}}:=\bm \Sigma(\beta \mathbf{F}^{T}{\bm y}+\bm D{\bm a})\label{eq:approx_para},
\end{equation}
and $\bm D$ is a diagonal matrix having diagonal elements $d_{1}^{-1},\dots,d_{N}^{-1}$.
We now need a way to fix the parameters ${\bm a, \bm d}$ of the prior. To do
so, we focus on the $n^{th}$ variable $w_{n}$ (with $1\leq n\leq N$),
and in particular on its approximated prior $\phi_{n}$. We can define
the \emph{tilted} distribution $Q^{(n)}$ as:
\begin{eqnarray}
Q^{(n)}({\bm w}|\mathbf{F},{\bm y}) & := & \frac{1}{Z_{Q^{(n)}}}e^{-\beta\frac{\left({\bm y}-\mathbf{F}\boldsymbol{w}\right)^{T}\cdot\left({\bm y}-\mathbf{F}\boldsymbol{w}\right)}{2}}\psi_{n}(w_{n})\prod_{l\ne n}\phi_{l}(w_{l};a_{l,}d_{l})\nonumber \\
 & = & \frac{1}{Z_{{Q}^{(n)}}}e^{-\frac{1}{2}({\bm w}-\bar{{\bm w}}_{(n)})^{T}\Sigma_{(n)}^{-1}({\bm w}-\bar{{\bm w}}_{(n)})}\psi_{n}(w_{n}),\label{eq:tilted_Q}
\end{eqnarray}
where:
\begin{equation}
\bm\Sigma_{(n)}^{-1}=\beta \mathbf{F}^{T}\mathbf{F}+\bm D_{(n)},\quad\bar{{\bm w}}_{(n)}=\bm\Sigma_{(n)}\left(\beta \mathbf{F}^{T}{\bm {\bm y}}+\bm D_{(n)}\boldsymbol{a}\right),\label{eq:tilted_para}
\end{equation}
and, in analogy with equation \eqref{eq:approx_para}, $\bm D_{(n)}$ is
a diagonal matrix of elements $d_{m}^{-1}$ for all diagonal elements
$m\neq n$ and zero for $m=n$. The \emph{tilted }distribution differs
from the approximated posterior $Q$ as it contains the original prior
$\psi_{n}$ instead of the $n$-th approximated prior
$\phi_{n}$. Intuitively, we expect that the ${n}$-th \emph{tilted}
distribution provides a better approximation of the expectation values
related to the $n$-th variable than the multivariate Gaussian
approximation. From a computational point of view,
the presence of a single intractable prior in
the \emph{tilted} distribution does not prejudice the efficiency of
the algorithm.

A natural way of fixing the optimal $\bm{a},\bm{d}$ parameters
consists in requiring the approximated $Q$ distribution to be as
similar as possible to the \emph{tilted} distribution $Q^{(n)}.$ To do
so, we minimize the Kullback-Leibler divergence
$D_{KL}\left(Q^{(n)}||Q\right)$. Perhaps not surprisingly, this
procedure is equivalent to equating the first two moments of the two
distributions:
\begin{eqnarray}
\left\langle w_{n}\right\rangle _{Q^{\left(n\right)}} & = & \left\langle w_{n}\right\rangle _{Q},\quad\left\langle w_{n}^{2}\right\rangle _{Q^{(n)}}=\left\langle w_{n}^{2}\right\rangle _{Q},\label{eq:mom_match}
\end{eqnarray}
where $\langle \cdot \rangle_Q$ and $\langle \cdot \rangle_{Q^{(n)}}$ denote averages w.r.t. $Q(\bm{w})$ and $Q^{(n)}(\bm{w})$, respectively.

Notice that the computation of the moments of the \emph{tilted}
distribution on the left-hand side of equation \eqref{eq:mom_match}
depends on the functional form of the prior considered.  We refer to
\ref{app:tilted_mom} for the expression of the moments of the tilted
distributions used in the case of a spike-and-slab prior.

Thanks to the multivariate Gaussian form of the approximated distribution,
it is a simple exercise to compute the moments of $Q$:
\begin{eqnarray}
\langle w_{n}\rangle_{Q}&=&\left(\frac{1}{d_{n}}+\frac{1}{\Sigma_{n,n}}\right)^{-1}\left(\frac{a_{n}}{d_{n}}+\frac{\bar{w}_{n}}{\Sigma_{n,n}}\right),\nonumber\\
\langle w_{n}^{2}\rangle_{Q}&=&\frac{1}{\frac{1}{d_{n}}+\frac{1}{\Sigma_{n,n}}}+\langle w_{n}\rangle_{Q}^{2}.\label{eq:approx_moms}
\end{eqnarray}
From equations \eqref{eq:approx_para} and \eqref{eq:tilted_para},
it is clear that the two matrices $\Sigma^{-1}$, and $\Sigma_{(n)}^{-1}$
differ only in a diagonal term. We can thus exploit a low-rank update
property to relate the two inverses. It turns out that the tilted
parameters are related to the approximated ones:
\begin{eqnarray}
\left(\bar{w}_{(n)}\right)_{n} & = & \left(\Sigma_{(n)}\right)_{n,n}\left(\frac{\bar{w}_{n}}{\Sigma_{n,n}}-\frac{a_{n}}{d_{n}}\right),\quad\left(\Sigma_{(n)}\right)_{n,n}=\frac{\Sigma_{n,n}}{1-\frac{\Sigma_{n,n}}{d_{n}}}.\label{eq:tilted_moms}
\end{eqnarray}
Upon imposing the moment matching condition \eqref{eq:mom_match}, we eventually
get an explicit equation for the prior parameters $a_{n},d_{n}$:
\begin{eqnarray}
d_{n} & = & \left(\frac{1}{\langle w_{n}^{2}\rangle_{Q^{(n)}}-\langle w_{n}\rangle_{Q^{(n)}}^{2}}-\frac{1}{\left(\Sigma_{(n)}\right)_{n,n}}\right)^{-1},\nonumber \\
a_{n} & = & \langle w_{n}\rangle_{Q^{(n)}}+\frac{d_{n}}{\left(\Sigma_{(n)}\right)_{n,n}}\left(\langle w_{n}\rangle_{Q^{(n)}}-\left(\bar{w}_{(n)}\right)_{n}\right).\label{eq:newad}
\end{eqnarray}
The ${\bm a},{\bm d}$ parameters are sequentially updated until a
fixed point is eventually reached: numerically, we need
to set a threshold below which the algorithm stops. To this purpose, for
each iteration $t$ (i.e. for each update of the ${\bm a},{\bm d}$
vectors), we can define an error $\epsilon_{t}$ as:
\[
\epsilon_{t}=\max_{n}\left|\left\langle w_{n}\right\rangle _{Q_{t}^{\left(n\right)}}-\left\langle w_{n}\right\rangle _{Q_{t-1}^{\left(n\right)}}\right|+\left|\left\langle w_{n}^{2}\right\rangle _{Q_{t}^{\left(n\right)}}-\left\langle w_{n}^{2}\right\rangle _{Q_{t-1}^{\left(n\right)}}\right|,
\]
where $Q_{t}^{\left(n\right)}$ is the tilted distribution with parameters
computed at iteration $t$. In practice, the algorithm stops
when $\epsilon_{t}<10^{-6}$. At convergence, the tilted distributions provide an approximation to the marginal densities of the posterior in equation \eqref{eq:posterior} and their first moments $\left\langle w_{n}\right\rangle _{Q^{\left(n\right)}}$ provide the estimate $\hat{\bm{w}}$ of the unknown vector $\bm{w}$.

For the sake of convenience, the EP procedure with low rank update that we have just presented is summarized in Algorithm \ref{EPlowrank}.

\begin{algorithm}
\caption{Expectation Propagation with low rank update}\label{EPlowrank}
\begin{algorithmic}[0]
\Procedure{EP}{$\mathbf{F}$, $\bm{y}$,$\{\psi_1,...,\psi_N\}$}
\State Initialize $\bm{a}^{old} $ and $\bm{d}^{old} $
\State $\mathbf{A}=\beta\mathbf{F}^T\mathbf{F}$
\Repeat
\State $\bm{\Sigma} =  \left(\mathbf{A}+\mathbf{D}\right)^{-1}$
\State $\bar{\bm{w}} = \bm{\Sigma}(\beta\mathbf{F}^T\bm{y}+\mathbf{D}\bm{a})$
\For {$k=1,...,N$}
\State Compute $\mu_k^{(k)}$ and $\Sigma_{kk}^{(k)}$ using the low rank update \eqref{eq:tilted_moms}.
\State Compute moments $\langle x_k \rangle_{Q^{(k)}}$ and $\langle x^2_k \rangle_{Q^{(k)}}$.
\State Compute $d_k^{new}$ and $a_k^{new}$ by moment matching using equation \eqref{eq:newad}.
\EndFor \Until convergence
\State \Return averages $\left\{\langle x_k \rangle_{Q^{(k)}}\right\}_{k=1,\dots,N}$ and variances $\left\{\langle x^2_k \rangle_{Q^{(k)}} - \langle x_k \rangle^2_{Q^{(k)}}\right\}_{k=1,\dots,N}$.
\EndProcedure
\end{algorithmic}
\end{algorithm}

\section{Experimental results with synthetic data}
\label{sec:exp}

In this section, we present our empirical results obtained by means of numerical simulations. We will first consider compressed sensing with i.i.d. random sensing matrices and then we will discuss some results related to the case of correlated random matrices.

\subsection{Uncorrelated measurements}

We consider $M\times N$ measurement matrices having i.i.d. entries
sampled from a standard normal distribution $\mathcal{N}(0,1)$. The
signal vector has $K=\rho N$ nonzero components, which are also
sampled from a Gaussian distribution with zero mean and unit variance.
The measurements are assumed to be noiseless. Note that, in general,
the parameter $\rho$ in equation \eqref{eq:l0} is unknown and needs to
be estimated. We show in \ref{app:rho_learn} how one can infer $\rho$
within the framework provided by EP.

We have run EP throughout the $\rho$-$\alpha$ plane. The parameters
used in our EP simulations are $\lambda=1$ and, in the finite
temperature formulation, $\beta=10^9$.

In order to measure the quality of the reconstruction, we consider the
sample Pearson correlation coefficient $r$ of the true vector and of the
reconstructed vector:
\begin{equation}
r=\frac{\sum_{k=1}^N (w_k - w_{sm})(\hat{w}_k-\hat{w}_{sm})}{\sqrt{\sum_{k=1}^N (w_k - w_{sm})^2} \sqrt{\sum_{k=1}^N (\hat{w}_k - {\hat{w}}_{sm})^2}},
\end{equation}
where $w_{sm}$ and $\hat{w}_{sm}$ are the sample means of the signal and of the inferred vector, respectively. We also consider the within-sample mean squared error as a measure of the
reconstruction error:
\begin{equation}
MSE = \frac{1}{N}\sum_{k=1}^{N}(w_k-\hat{w}_k)^2.
\end{equation}

\begin{figure}%
\centering
\subfigure[$N=200$]{
\includegraphics[width=0.47\textwidth]{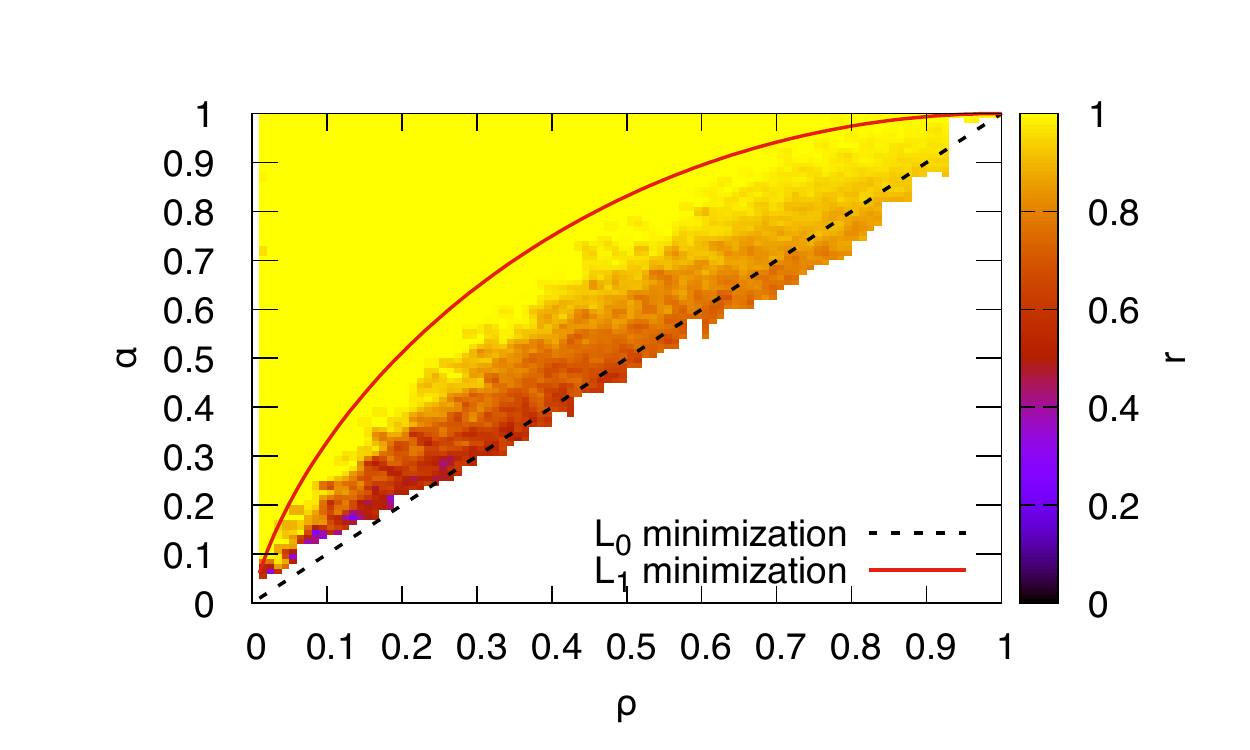}\label{phdiag200}}
\quad
\subfigure[$N=400$]{
\includegraphics[width=0.47\textwidth]{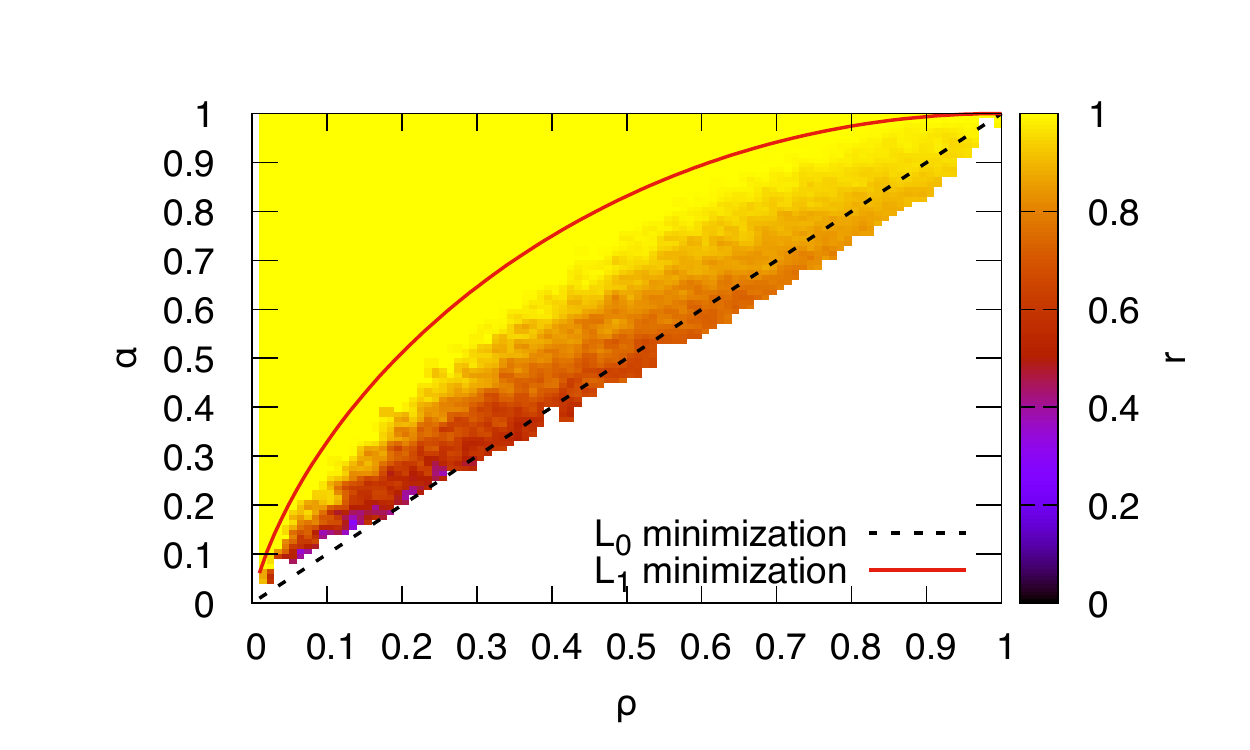}\label{phdiag400}}
\\
\subfigure[$N=800$]{
\includegraphics[width=0.47\textwidth]{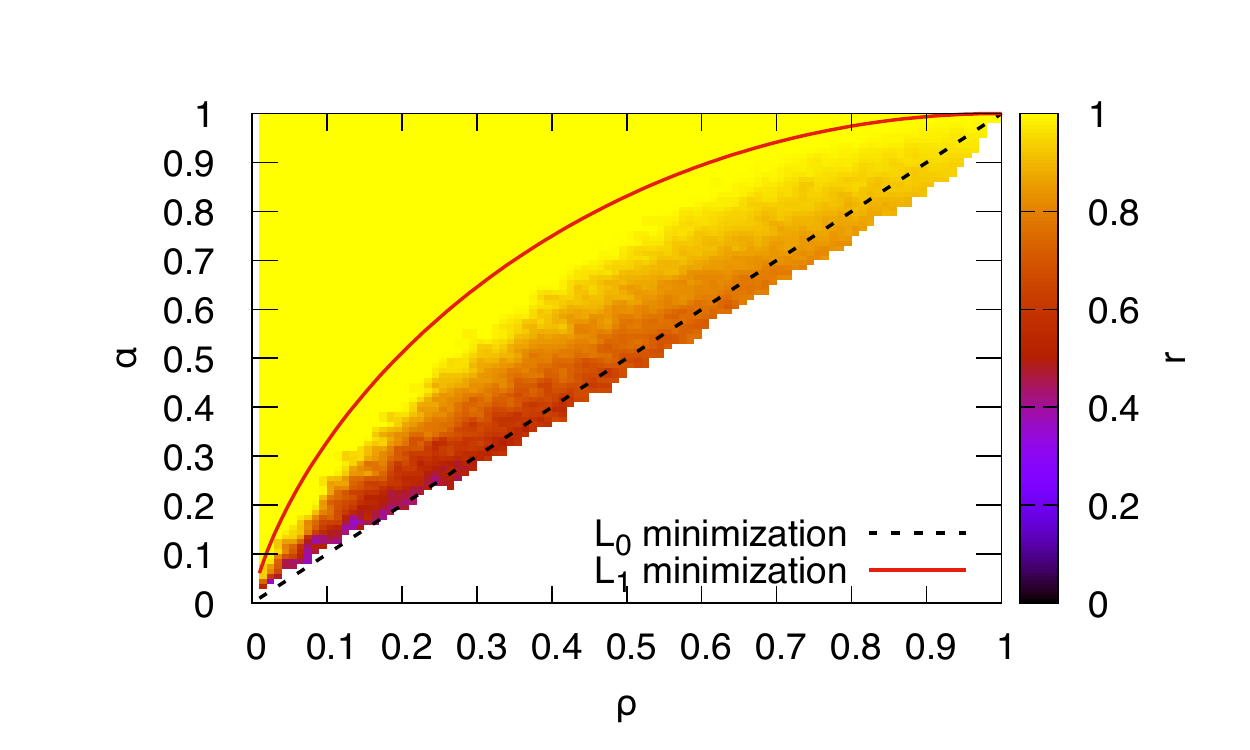}\label{phdiag800}}
\quad
\subfigure[$N=1600$]{
\includegraphics[width=0.47\textwidth]{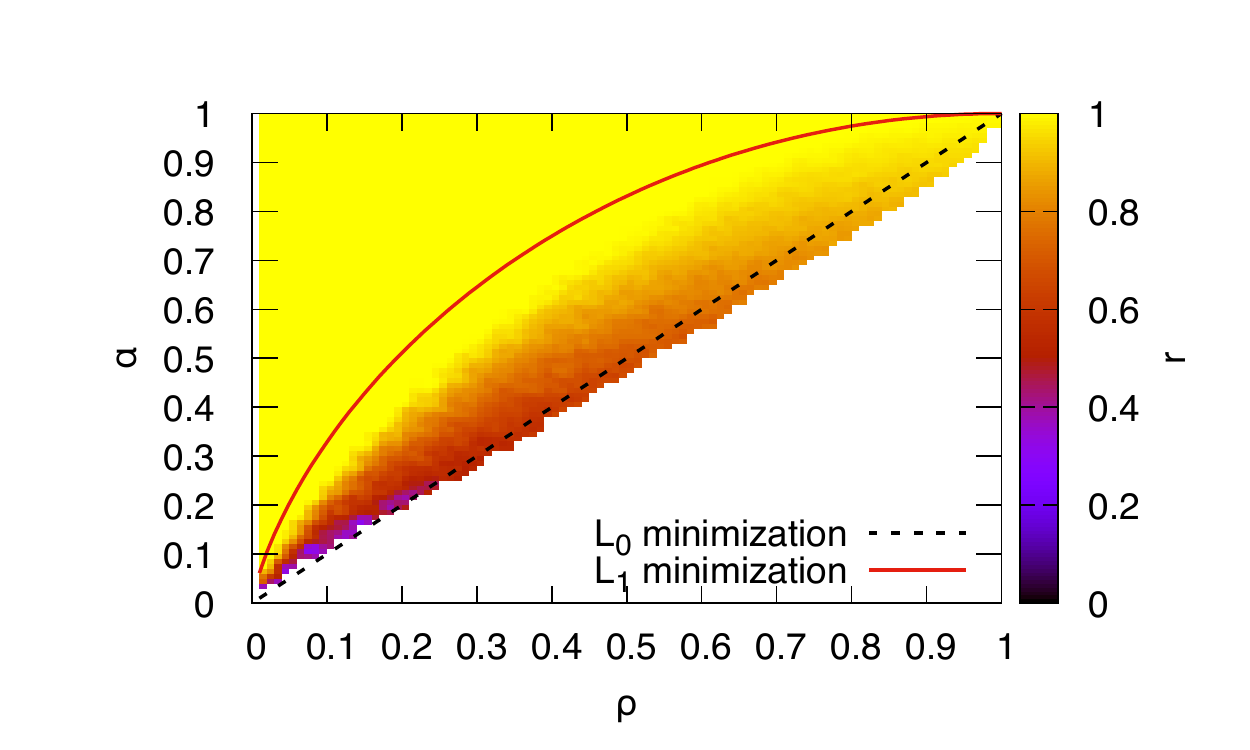}\label{phdiag1600}}
\caption{Compressed sensing phase diagram for $N=200,400,800,1600$. Each point corresponds to a single simulation. The color refers to the Pearson correlation coefficient between the true vector and the reconstructed vector. We plot the lines corresponding to the $L_1$ reconstruction and to the theoretical reconstruction limit, given by the $L_0$ condition.}
\label{phdiag}
\end{figure}

We plot the sample correlation coefficient in figure \ref{phdiag} for a single simulation at each given $\rho$ and $\alpha$ and progressively larger values of $N$, namely $N=200$ (figure \ref{phdiag200}), $N=400$ (figure \ref{phdiag400}), $N=800$ (figure \ref{phdiag800}) and $N=1600$ (figure \ref{phdiag1600}). The $L_{0}$ line represents the theoretical limit $M = K$, under which a perfect reconstruction is impossible, whereas the $L_1$ line was obtained in \cite{kabashima,erratumkabashima} using the replica method in the limit $N\rightarrow\infty$ and $M\rightarrow\infty$, with $\alpha$ finite. The white region is the one in which EP does not converge.

\begin{figure}
\centering
\includegraphics[scale=0.4]{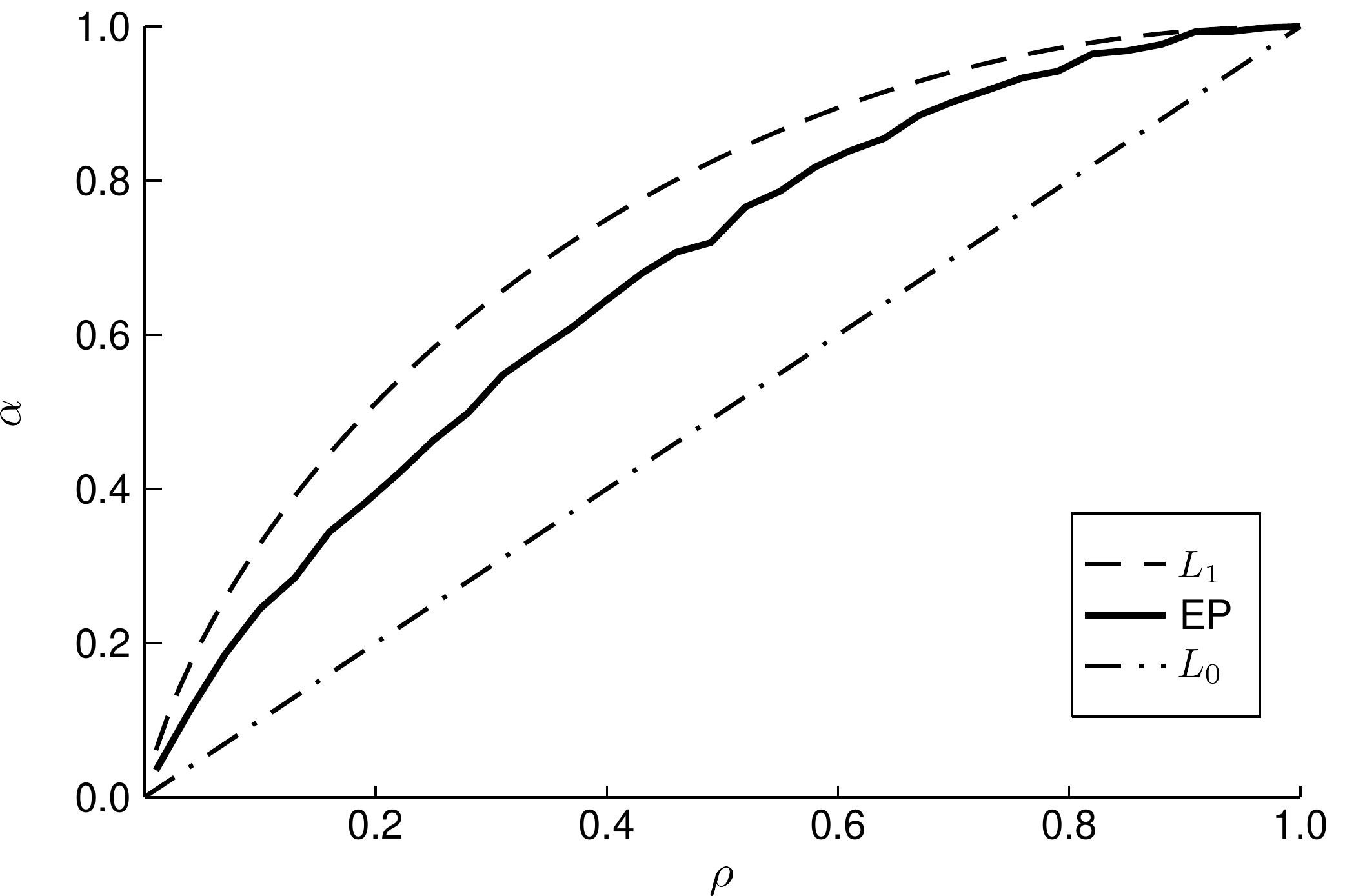}
\caption{EP phase transition line as obtained using the bisection-like algorithm described in the main text. The number of variables is $N=1600$ and the threshold $\delta$ for the difference of the mean squared error of the evaluated points is $10^{-5}$.}
\label{EPline}
\end{figure}

The plots suggest that there exists a phase transition line $\{(\rho,\alpha^{EP}(\rho)), 0\leq\rho\leq 1\}$, which is located under the $L_1$ line. In order to obtain the coordinates of the points along the transition line, one can proceed numerically by using a bisection-like algorithm.
After discretizing the interval $0\leq \rho \leq 1$, one can select two starting values of $\alpha$ for each discretized value $\rho_0$. For instance, a reasonable choice, which is the one we adopt, is taking $\alpha_0$ on the $L_0$ minimization line, namely $\alpha_0(\rho_0) = \rho_0$, and $\alpha_1(\rho_0)$ on the $L_1$ minimization line, where $\alpha_1(\rho_0)$ is expressed as \cite{kabashima}:
\begin{equation}
\alpha_1(\rho_0)=2(1-\rho_0)H(\hat{\chi}^{-1/2})+\rho_0.
\end{equation}
In the last equation, $\hat{\chi}$ is given by the solution of \cite{erratumkabashima}:
\begin{equation}
\hat{\chi}=\alpha^{-1}\left[2(1-\rho_0)\left((\hat{\chi}+1)H(\hat{\chi}^{-1/2})-\hat{\chi}^{1/2}\frac{e^{-1/(2\hat{\chi})}}{\sqrt{2\pi}}\right)+\rho_0(\hat{\chi}+1)\right],
\end{equation}
where $H(x)=\int_x^{+\infty}\exp(-\frac{t^2}{2})/(2\pi)dt$. Then, one performs the EP inference for configurations corresponding to those points and to the point $(\rho^*, \alpha^*)$, where $\rho^*=\rho_0$ and $\alpha^*=(\alpha_0+\alpha_1)/2$ and computes their mean squared error. If the difference $|MSE(\alpha_1)-MSE(\alpha^*)|$ is negligible (i. e. smaller than a certain threshold $\delta$), then we set $\alpha_1 = \alpha^*$. Otherwise, we set $\alpha_0 = \alpha^*$. We recompute the middle point and repeat the procedure until we reach the desired accuracy $\Delta\alpha_{min}$ on the points located at the boundary between the successful and unsuccessful reconstruction regions. We summarize the procedure in Algorithm \ref{bisection_alg}. The resulting transition line is shown in figure \ref{EPline}.

\begin{algorithm}
\caption{Bisection algorithm}\label{bisection_alg}
\begin{algorithmic}[0]
\Procedure{bisection}{$N,\rho_0$,$\alpha_0$,$\alpha_1$;$\delta$,$\Delta\alpha_{min}$}
\State Set $K=\rho_0 N$.
\State Set $\alpha^* = (\alpha_0+\alpha_1)/2$.
\Repeat
\State Set $M_1 = \alpha_1 N$.
\State Generate signal $\bm w_1$ and sensing matrix $\mathbf{F}_1$.
\State Infer $\hat{\bm{w}}_1$ using Algorithm \ref{EPlowrank} with inputs $\bm{y}_1=\mathbf{F}_1\bm{w}_1$, $\mathbf{F}_1$ and $\rho=\rho_0$.
\State Compute $MSE(\alpha_1)$ between $\hat{\bm{w}}_1$ and $\bm{w}_1$.
\State Set $M^* = \alpha^* N.$
\State Generate signal $\bm w^*$ and sensing matrix $\mathbf{F}^*$.
\State Infer $\hat{\bm{w}}^*$ using Algorithm \ref{EPlowrank} with inputs $\bm{y}^*=\mathbf{F}^*\bm{w}^*$, $\mathbf{F}^*$ and $\rho=\rho_0$.
\State Compute $MSE(\alpha^*)$ between $\hat{\bm{w}}^*$ and $\bm{w}^*$.
\If{$|MSE(\alpha_1) - MSE(\alpha^*)|>\delta$}
\State $\alpha_0 = \alpha^*$.
\Else
\State $\alpha_1 = \alpha^* $.
\EndIf
\State Reassign $\alpha^* = (\alpha_0+\alpha_1)/2$.
\Until $|\alpha_1-\alpha_0|/2<\Delta\alpha_{min}$
\State \Return $\alpha^*$
\EndProcedure
\end{algorithmic}
\end{algorithm}

\begin{figure}
\centering
\subfigure[]{
\includegraphics[width=0.42\textwidth]{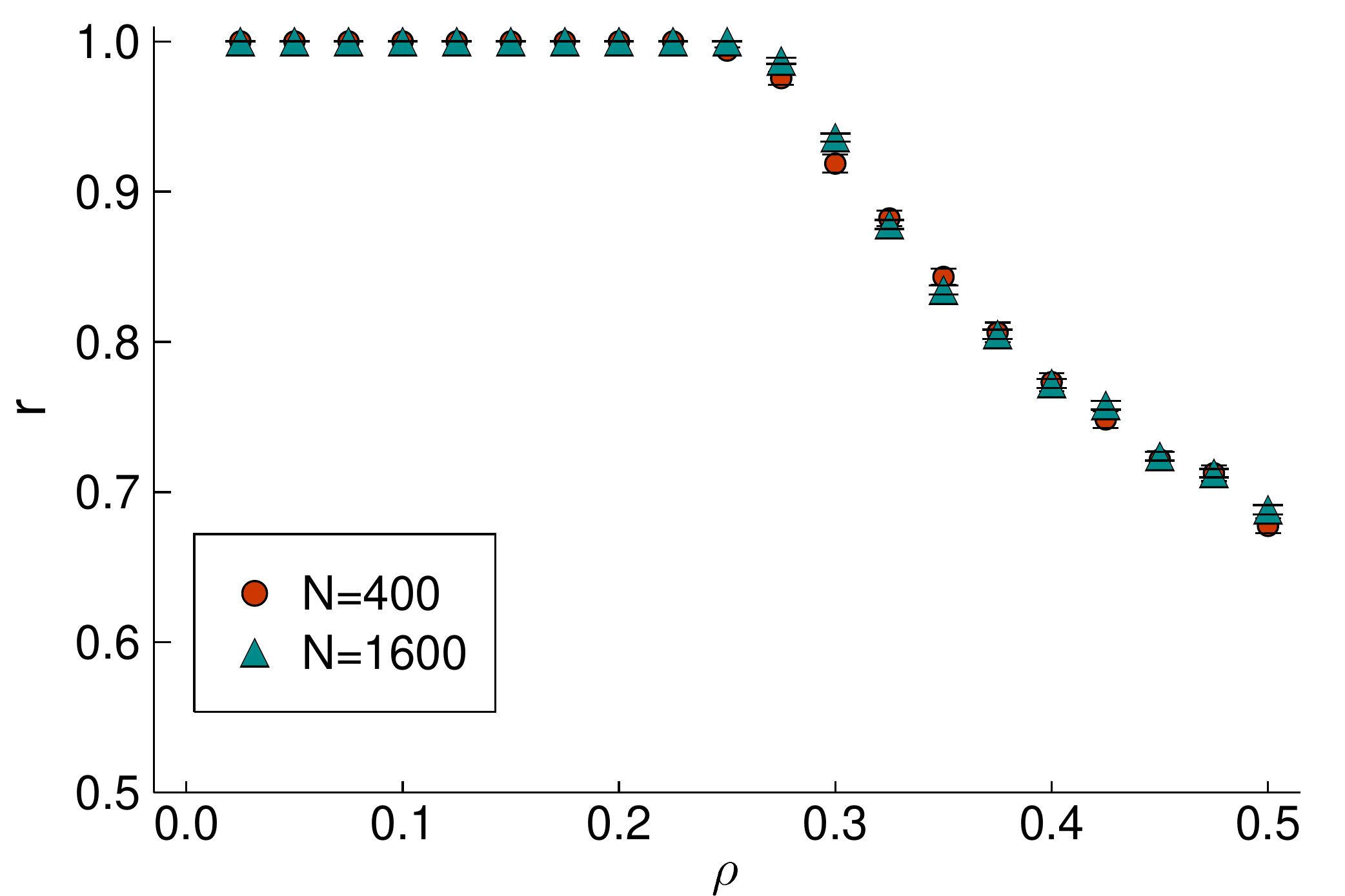}}
\quad
\subfigure[]{
\includegraphics[width=0.45\textwidth]{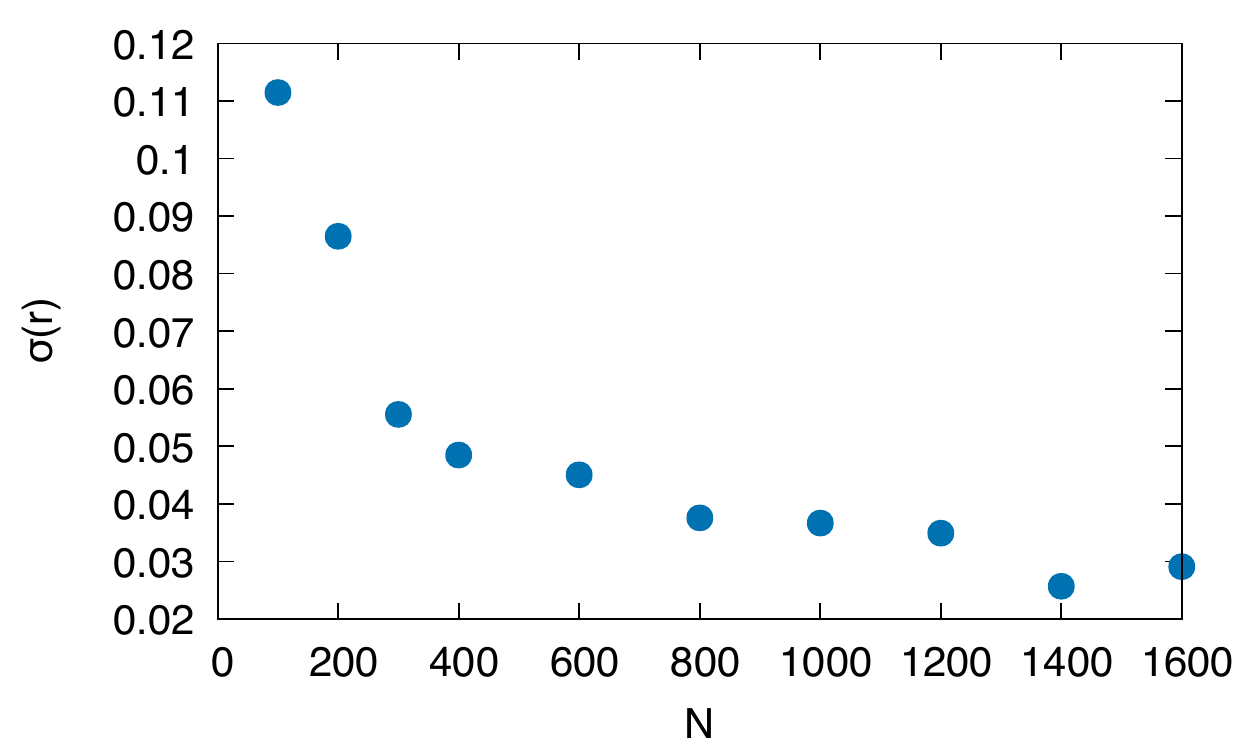}}
\caption{(a) Pearson correlation as a function of $\rho$ at $\alpha = 0.5$, for $N=400$ and $N=1600$. The error bars are estimated as $\sigma(r)/N_t$, where $N_t=100$ is the number of trials. (b) Sample standard deviation of $r$ as a function of $N$. The value of the density and of the measurement rate are fixed and given by $\rho=0.4$ and $\alpha=0.55$.}
\label{r_fluctuations}
\end{figure}

We define {\it probability of convergence} as the empirical frequency that for a
random instance of the signal $\bm{w}$ and of the measurement matrix
$\mathbf{F}$, the algorithmic error $\epsilon_t$ becomes arbitrarily small after
some iteration, as the maximum number of iterations $t$ is increased. In
practice, we  set a threshold for the error $\epsilon_t$ equal to $10^{-6}$ and
we  estimated the probability of convergence as the fraction of times the
algorithm fulfilled the convergence criterion. Empirically, it turns out that
the probability of convergence of EP,  increases with the number of variables
$N$ (not shown).  Moreover, the fluctuations of the Pearson correlation
coefficient $r$ and of the MSE beyond the transition line decrease as the number
of variables $N$ becomes larger, whereas their average values do not seem to
depend on the size of the system. We show this in the case of the MSE in figure
\ref{r_fluctuations}, for $N=400$ and $N=1600$.

\begin{figure}
\centering
\subfigure[]{
\includegraphics[width=0.45\textwidth]{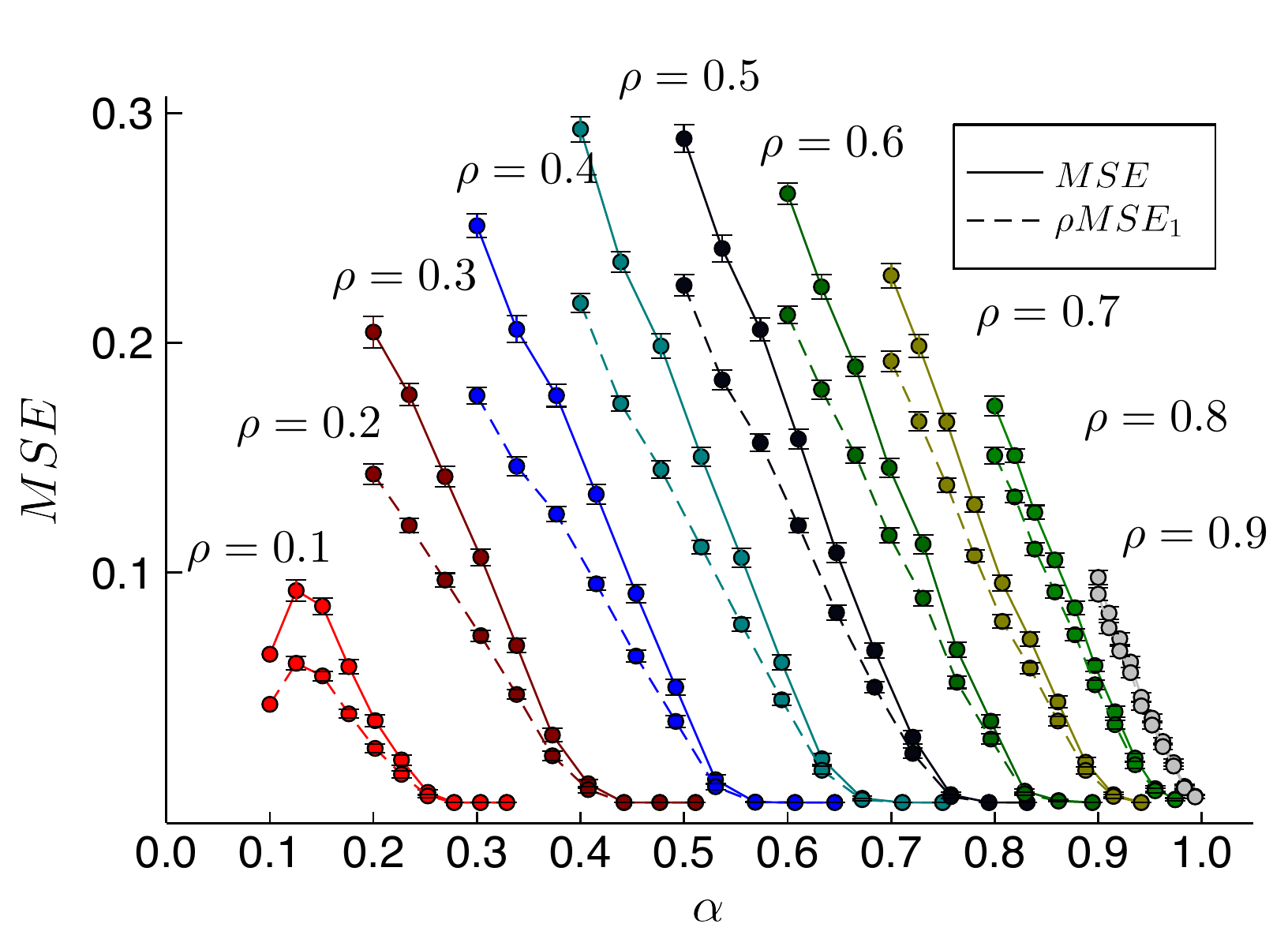}
\label{MSEhead}}
\quad
\subfigure[]{
\includegraphics[width=0.45\textwidth]{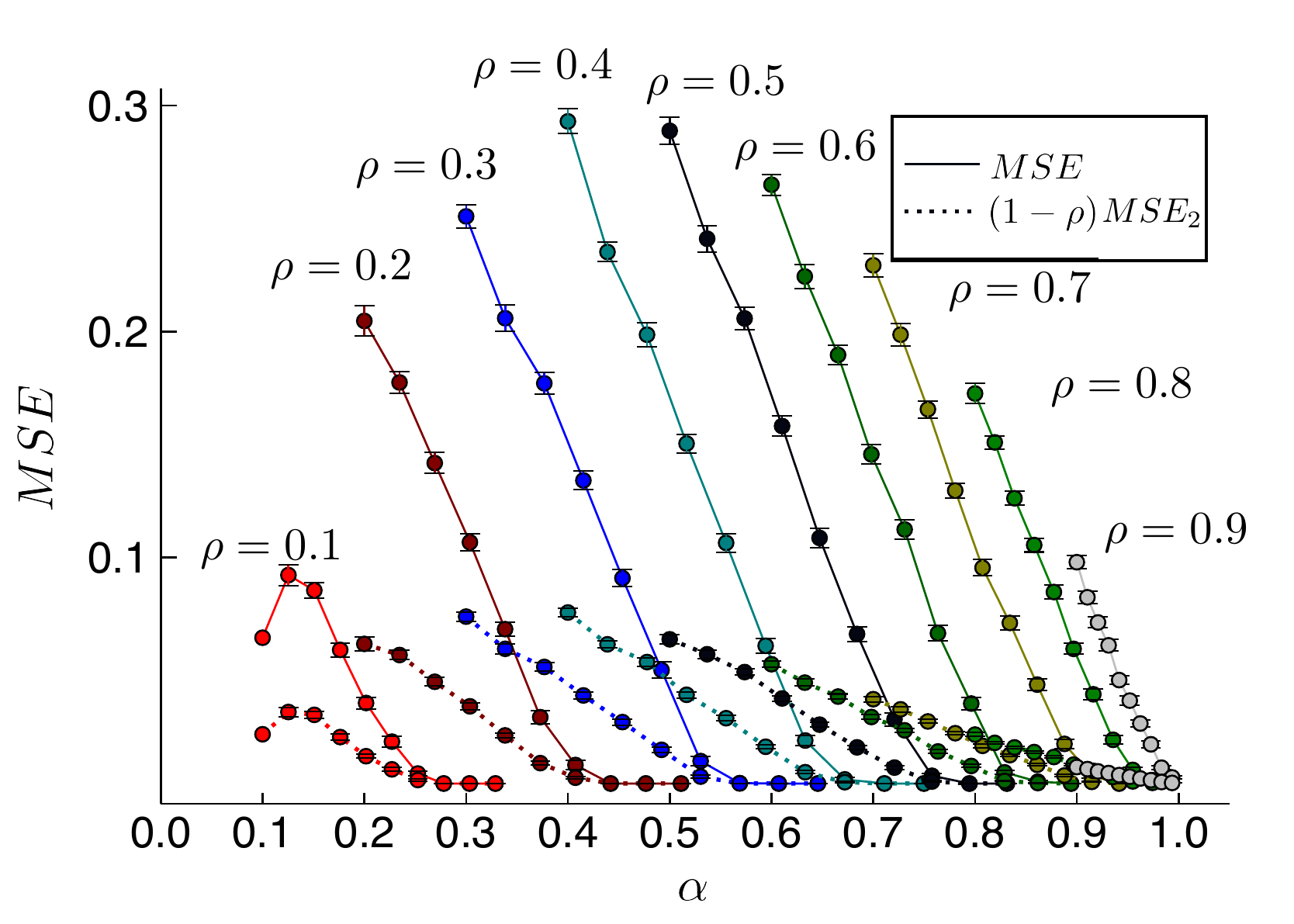}
\label{MSEtail}}
\caption{a) Contribution of the first $K$ components to the mean squared error (dashed lines), compared to the mean squared error itself (solid lines). (b) Contribution of the last $N-K$ components (the `tail' of the vector) to the mean squared error (dotted lines), compared to the mean squared error itself (solid lines).
In all plots, $N=400$, the number of simulations is 100 and each curve corresponds to a different value of $\rho$. The points are averages computed over the $N_c$ converged simulations and the error is estimated from the sample standard deviation $\sigma$  as $\sigma/\sqrt{N_c}$. From left to right, $\rho$ ranges from $0.1$ to $0.9$.}
\label{weight_of_tail}
\end{figure}

Finally, we note that the mean squared error can be expressed as follows:
\begin{equation}
MSE = \rho MSE_1 + (1-\rho) MSE_2,
\end{equation}
where $MSE_1$ is the mean squared error between the vector of the $K$ non zero components of the signal and the corresponding vector extracted from the inferred signal and $MSE_2$ is the mean squared error between the original and inferred vectors having the remaining $N-K$ elements as components. The latter corresponds to the squared norm of the last $N-K$ components of the inferred vector (divided by $N-K$). Beyond the CS threshold of EP, the dominant contribution to the reconstruction error comes from the estimate of the $K$ non-zero components of the reconstructed vector, implying that, overall, EP is still quite accurate in discriminating the zero entries of the signal. This is shown in figures \ref{MSEhead} and \ref{MSEtail}, in which, respectively, $\rho MSE_1$ and $(1-\rho)MSE_2$ are compared to the total mean squared error.

\subsection{Correlated measurement matrices}
We consider the case of correlated measurement matrices $\mathbf{F}$:
\begin{equation}
\mathbf{F} = ({\bm{f}}_1,
...,
{\bm{f}}_M)^T,
\label{eq:cor_mat1}
\end{equation}
whose rows $\bm{f}_i\in\mathbb{R}^N$ are correlated but linearly independent samples drawn from a multivariate Gaussian distribution:
\begin{equation}
\bm{f}_i \sim \mathcal{N}(\bm{0},\mathbf{S}), \qquad i=1,...,M.
\label{eq:cor_mat2}
\end{equation}
The covariance matrix $\mathbf{S}$ is designed according to the following functional form:
\begin{equation}
\mathbf{S} = \mathbf{Y}^T\mathbf{Y}+\bm{\Delta},
\label{eq:cov_mat}
\end{equation}
where $\mathbf{Y}$ is a $k\times N$ matrix with random i.i.d. Gaussian $\mathcal{N}(0,1)$ entries and having a controllable rank $k$ and $\bm{\Delta}$ is a diagonal $N\times N$ matrix with positive Gaussian i.i.d. eigenvalues.
 Notice that the product $\mathbf{Y}^T\mathbf{Y}$ is symmetric and positive semi-definite by construction. Adding the matrix $\bm{\Delta}$ ensures that $\mathbf{S}$ has maximum rank.

\begin{figure}
\centering
\subfigure[]{
\includegraphics[width=0.45\textwidth]{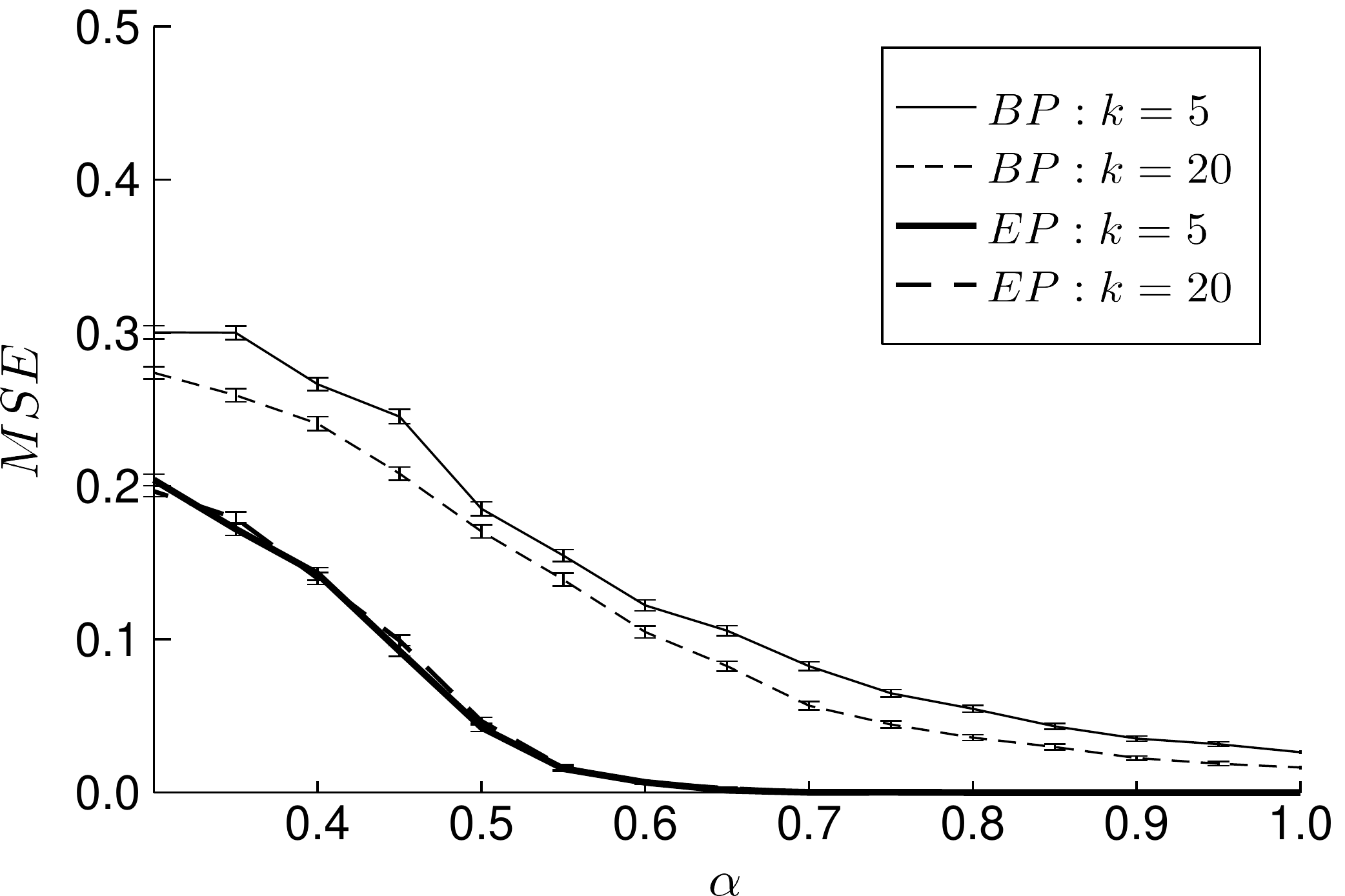}
\label{MSE-EPBP-RHO3e-1}}
\quad
\subfigure[]{
\includegraphics[width=0.45\textwidth]{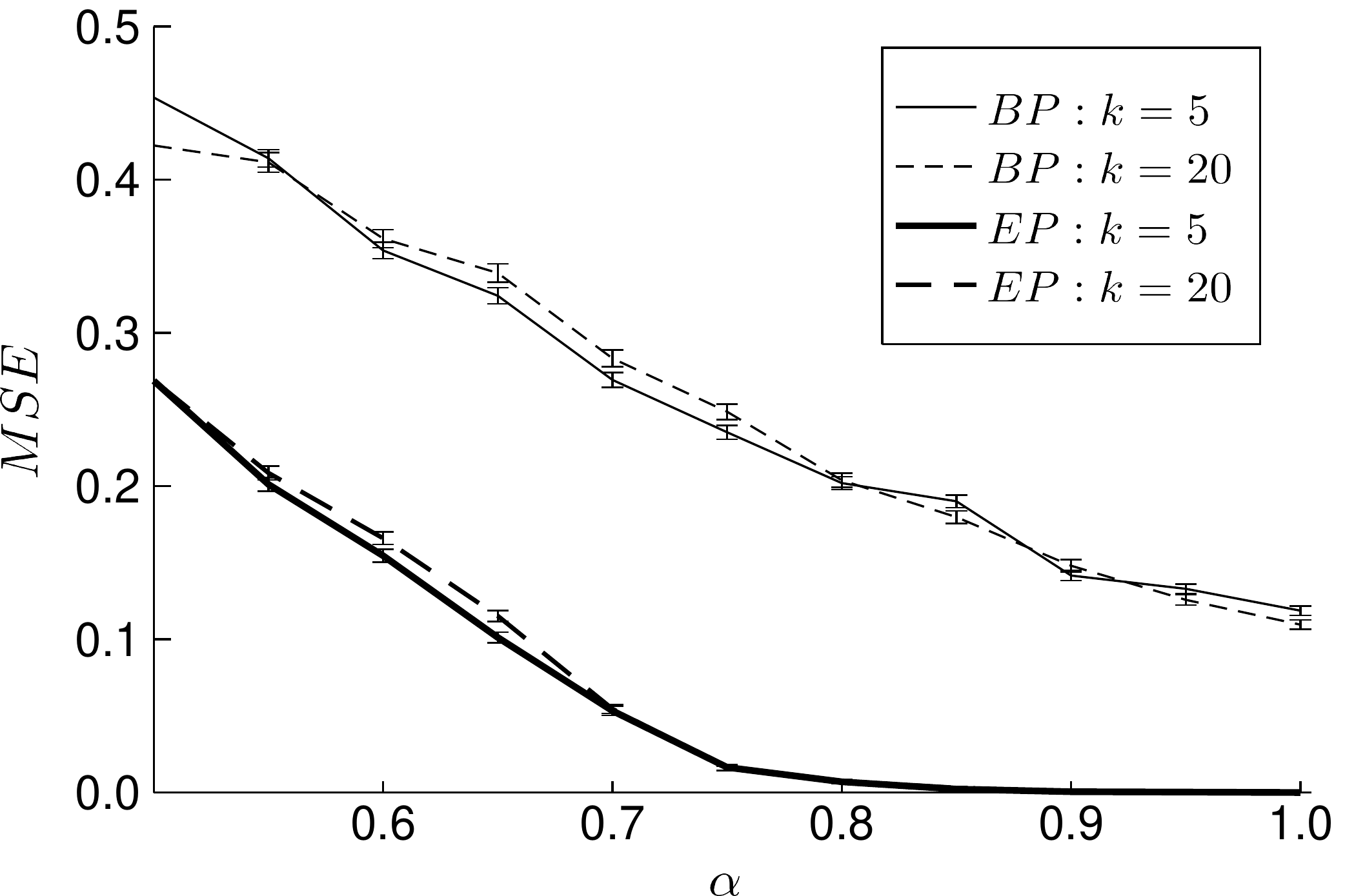}
\label{MSE-EPBP-RHO5e-1}}\\
\subfigure[]{
\includegraphics[width=0.45\textwidth]{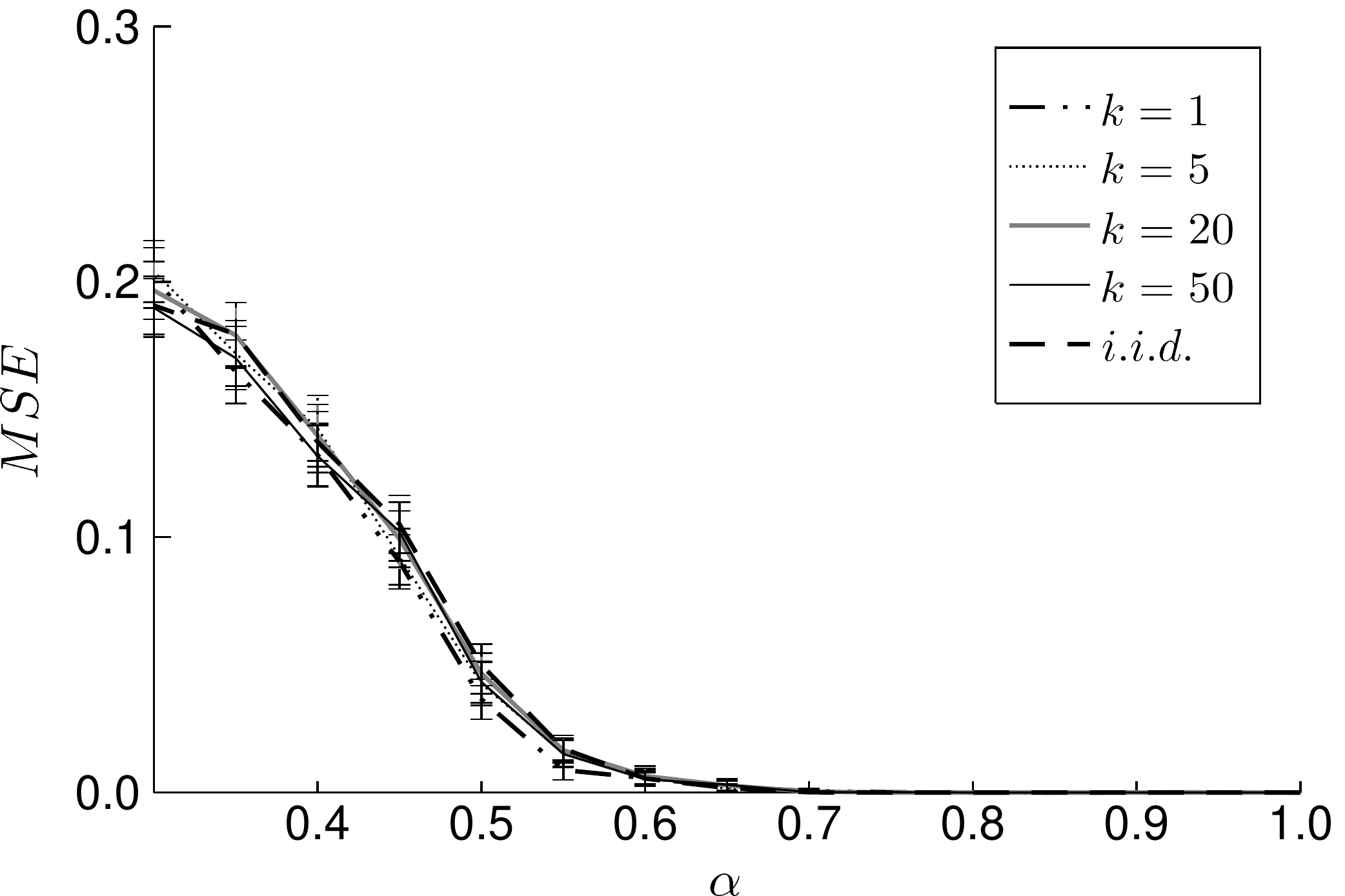}
\label{MSE-EP-RHO3e-1}}
\quad
\subfigure[]{
\includegraphics[width=0.45\textwidth]{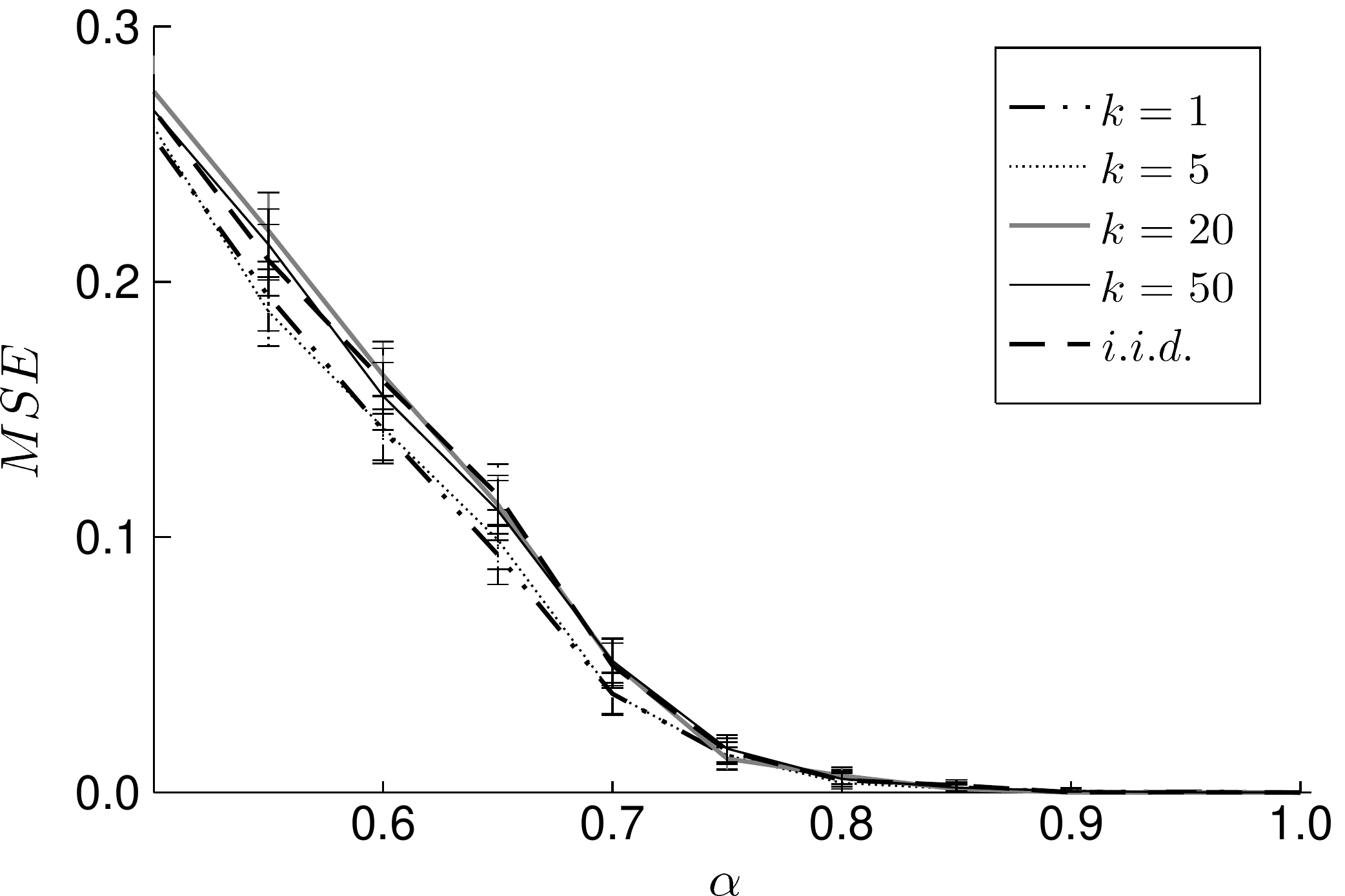}
\label{MSE-EP-RHO5e-1}}
\caption{(a),(b) Comparison between the MSE obtained when reconstructing by means of EP and by means of EMBP in the case of correlated measurement matrices. The plot shows the MSE as evaluated over $N_t=1000$ converged trials and the uncertainty has been estimated as $\sigma/\sqrt{N_t}$, where $\sigma$ is the sample standard deviation. Regardless of the value of $\alpha$, the EMBP algorithm is not able to reconstruct correctly the signal, whereas EP achieves zero MSE beyond a critical value $\alpha_c(\rho)$.
(c),(d) MSE resulting from the reconstruction from both i.i.d. and correlated measurement matrices. Each point was evaluated over $N_t=1000$ trials. Lower values of $k$ correspond to more correlated measurements.
The number of variables is $N=50$ and the density of the signal is (a),(c) $\rho = 0.3$ and (b),(d) $\rho=0.5$.
}
\label{MSEcorrelated}
\end{figure}

We first study the retrieval performance of EP and of Expectation Maximization Belief Propagation (EMBP), a similar message passing reconstruction algorithm \cite{PhysRevX.2.021005,krzakala}, implemented in MATLAB and available at \texttt{http://aspics.krzakala.org}.

The BP approximation lies on the independence of
the entries of the sensing matrix, a condition that is not generally
fulfilled in the matrices considered in this section. In particular,
for small values of $k$ the covariances and the variances of
$\mathbf{S}$ are of the same order of magnitude. However, as $k$
increases, these variances become dominant with respect to the
off-diagonal entries of $\mathbf{S}$ and the associated multivariate
Gaussian measure becomes more and more similar to the product of
independent univariate Gaussian distributions.  In figures
\ref{MSE-EPBP-RHO3e-1} and \ref{MSE-EPBP-RHO5e-1}, we compare the MSE
associated with EMBP and with EP when using correlated matrices and
for $\rho=0.3$, $\rho=0.5$ respectively. The signal density is learned
in both cases, using Expectation Maximization in the case of EMBP and
using the free energy-based variational method described in
\ref{app:rho_learn} in the case of EP.
Only converged trials have been taken into account. While BP fails to
correctly reconstruct the signal and to infer the signal density in
the presence of the correlations we introduced in the measurements,
the performance of the EP reconstruction is unaffected. As we
expected, EMBP performances improve as $k$ increases. However, we note
that at low enough values of $k$, such as those we have considered,
EMBP never achieves zero MSE, not even when $\alpha=1$ (that is when
we have as many equations as variables).

The fraction of converged trials is generally far lower in the case of
EMBP than in the case of EP and decreases as $N$ is increased.  For
example, in the case of $k=1$, for $N=50$, it is of the order of one
in a thousand of simulated trials, two orders of magnitude lower than
the fraction of converged EP simulations at the same values of $N$ and
$k$ (not shown).

We also plot the MSE resulting from the EP reconstruction for i.i.d. measurement matrices and correlated measurement matrices that are constructed using equations \eqref{eq:cor_mat1}, \eqref{eq:cor_mat2} and \eqref{eq:cov_mat} for various values of $k$.
By considering $k=50$, $k=20$, $k=10$, $k=5$ and $k=1$, we obtain that the associated mean squared errors do not exhibit significant discrepancies, as shown in figures  \ref{MSE-EP-RHO3e-1} and \ref{MSE-EP-RHO5e-1}.
This confirms that the EP inference of the signal is not altered with respect to the case of i.i.d. sensing matrices and retains its correct-incorrect reconstruction threshold.

Finally, we tested several algorithms for sparse reconstruction on linear systems with the same type of correlated measurement matrices considered so far. More precisely, these algorithms are Basis Pursuit \cite{C05_l1-magic:recovery}, Orthogonal Matching Pursuit (OMP) \cite{OMP}, Regularized Orthogonal Matching Pursuit \cite{ROMP}, Compressive Sampling Matching Pursuit (CoSaMP) \cite{NEEDELL2009301}, Subspace Pursuit \cite{4839056}, Smoothed L0 (SL0) \cite{4663911}, Approximate Message Passing (AMP) \cite{donoho2009mp} and, again, Expectation Maximization Belief Propagation \cite{PhysRevX.2.021005}.
These algorithms are implemented in the C++ library KL1p \cite{kl1p}. This specific implementation makes use of the linear algebra library Armadillo \cite{armadillo_2018, armadillo_2016}.

\begin{figure}
\centering
\subfigure[]{
\includegraphics[width=0.7\textwidth]{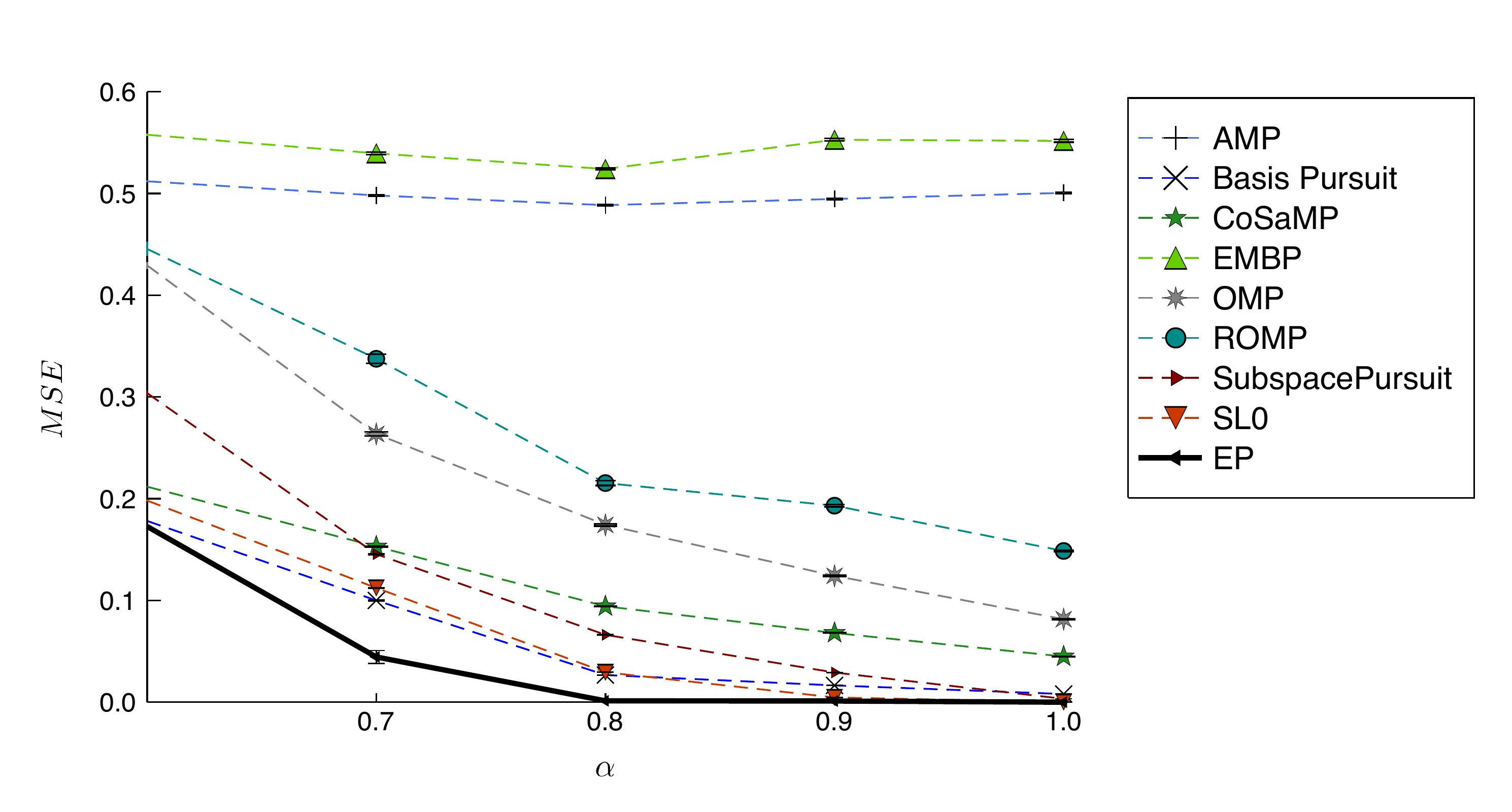}
\label{MSE-ALL}}
\\
\subfigure[]{
\includegraphics[width=0.35\textwidth]{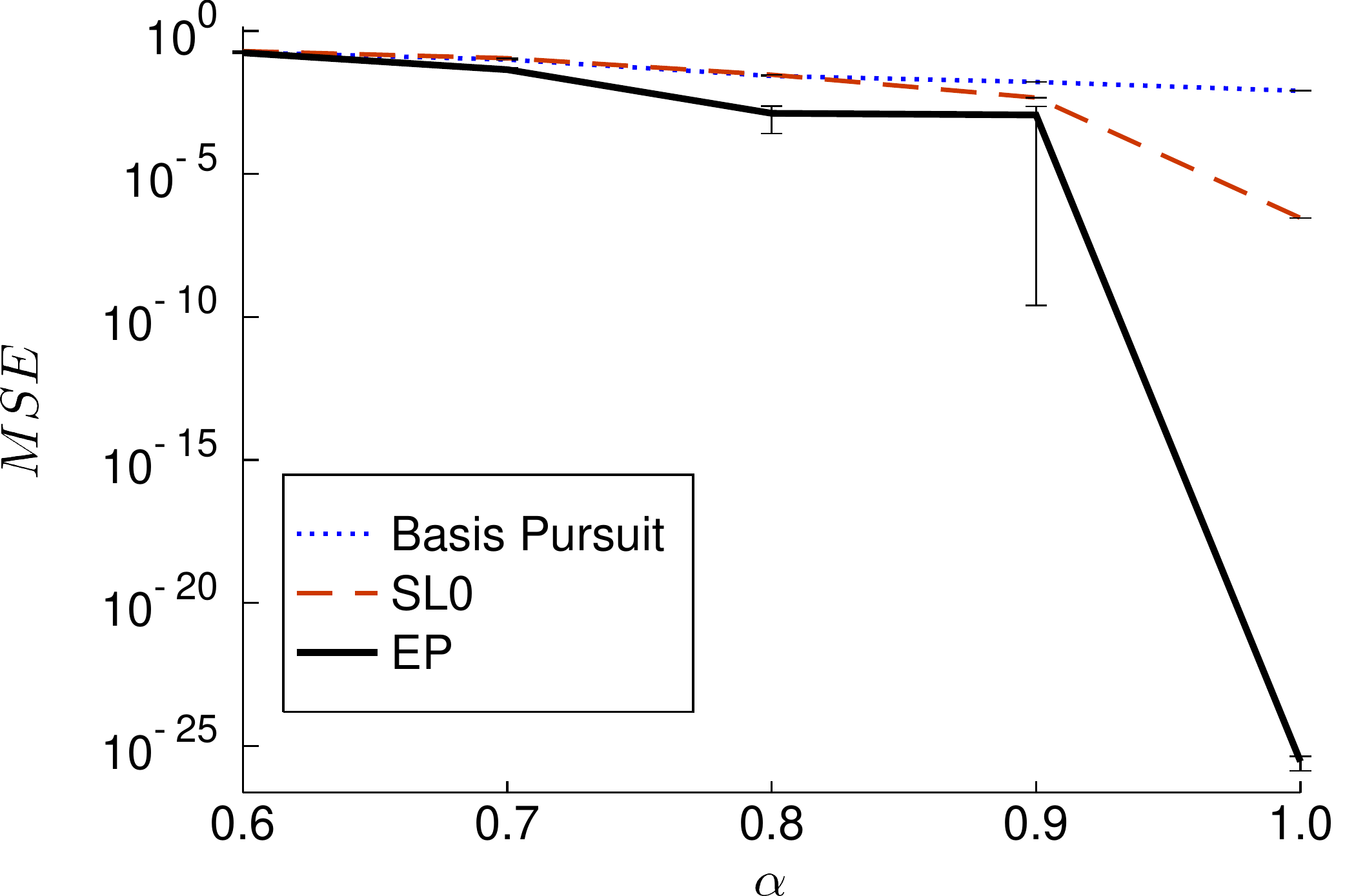}
\label{MSE-ALL-LOG}}
\quad
\subfigure[]{
\includegraphics[width=0.53\textwidth]{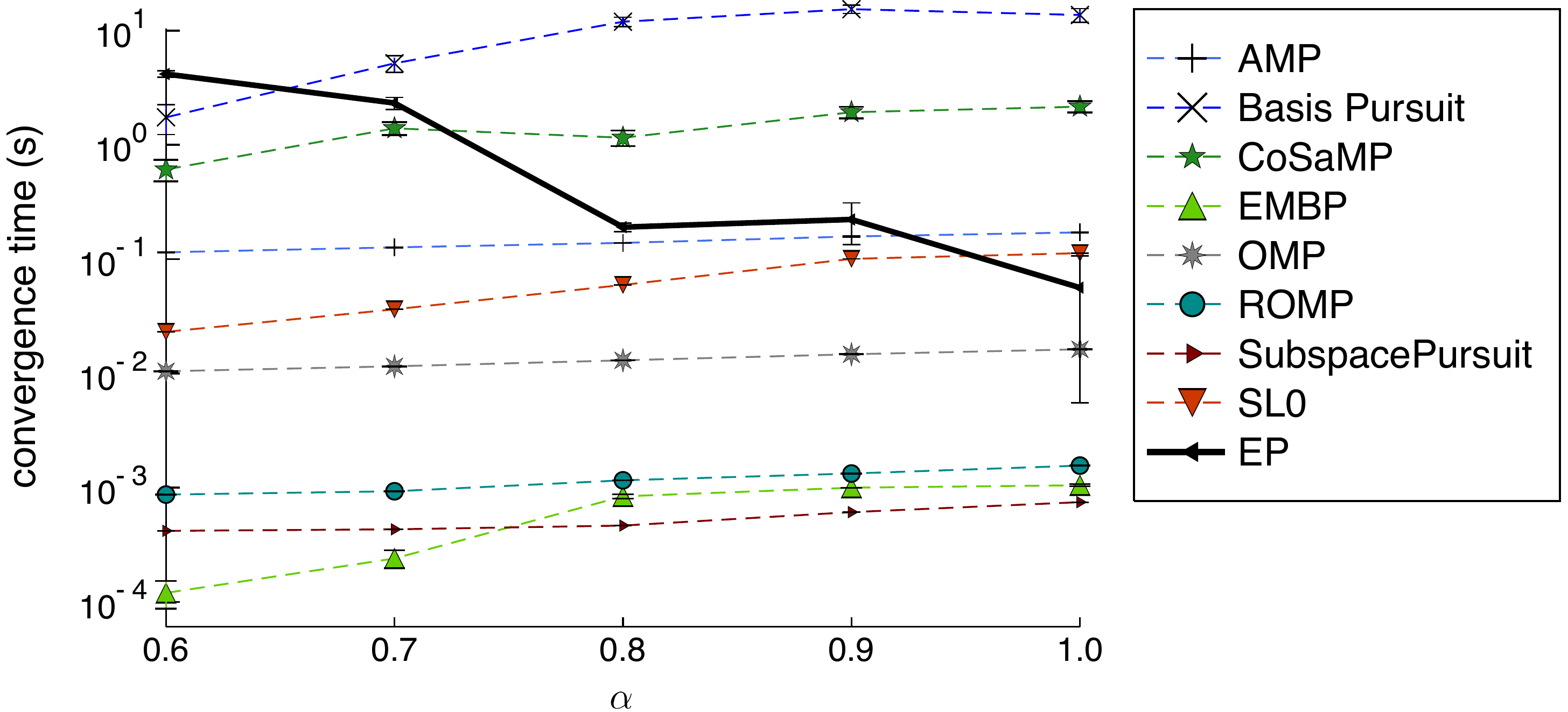}
\label{TIMES-ALL}}
\caption{
(a) Comparison of the MSE of various CS reconstruction algorithms in the presence of correlated measurement matrices with $k=5$. The signals to be retrieved have density $\rho=0.5$. (b) Semi-logarithmic plot of the reconstruction error of Basis Pursuit, SL0 and EP. (c) Comparison of the elapsed running time of the same reconstruction algorithms in the presence of correlated measurement matrices. In both figures, the parameters of the generated signals are given by $N=100$, $\rho=0.5$ and $\kappa$=5 and the total number of trials is $N_t=100$.}
\label{fig:ALL-ALGORITHMS}
\end{figure}

In order to compare the performance of these algorithms, we generated $N_t=100$ random gaussian i.i.d. signals of length $N=100$ and as many random correlated sensing matrices, with $k=5$.
For any given pair of signal $\bm w$ and measurement matrix $\mathbf F$, we attempted to recover the original signal by means of EP and of the algorithms included in KL1p.
The results are presented in figure \ref{fig:ALL-ALGORITHMS}. As we can see in figure \ref{MSE-ALL} and as further highlighted in the semi-logarithmic plot in figure \ref{MSE-ALL-LOG}, EP is the only algorithm exhibiting an incorrect-correct reconstruction phase transition, whereas all the other methods that we considered fail to retrieve the signal regardless of the value of $\alpha$.
In terms of running time, EP appears to be comparable to most of the other reconstruction techniques, as shown in figure \ref{TIMES-ALL}.

\section{Conclusions}
\label{sec:concl}
We have proposed an EP-based scheme for efficient CS reconstruction
whose computational complexity is dominated by a matrix inversion per
iteration which requires $O(N^3)$ operations

By analyzing the reconstruction achieved by EP in the case of
undersampled linear systems with random i.i.d. measurement matrices,
we showed that EP exhibits a phase transition, in analogy to
other message passing inspired algorithms.  We numerically computed
the threshold and the related phase transition diagram and found that
signal reconstruction is possible below the $L_1$ minimization line
(see figures \ref{phdiag}, and \ref{EPline}).

Finally, we investigated the case of correlated measurement matrices
and found that the EP threshold persists, implying that EP is still
capable of accurately retrieving the signal beyond a critical $\alpha$
and that, contrary to the case of other reconstruction algorithms, it
is robust against the presence of statistical structure in the
measurements.

\section{Acknowledgements}
We acknowledge the kind hospitality of the Physics Department of the
University of Havana where part of this work has been completed.  This
work has received support by INFERNET -- Marie Sk{\l}odowska-Curie grant
agreement number 734439 -- and by the SmartData@PoliTO interdepartmental center of Politecnico di Torino.

\appendix

\section{Expectation Propagation in the zero temperature limit}
\label{app:T=0}

In section \ref{sec:ep}, we implemented the linear constraints in
equation \eqref{eq:y=Aw} as the $\beta\rightarrow\infty$ limit
of the multivariate Gaussian measure. We are going to show here how
one can compute analytically this limit. Let us rewrite the previous
formalism in a slightly different matrix notation. We will assume that the rank of the measurement matrix is maximum and equal to $M$, with $M<N$
(if this is not the case, we can easily remove the linearly dependent
rows from the measurement matrix $\mathbf{F}$). Through Gaussian elimination,
we can transform the matrix $\mathbf{F}$ to a row echelon form:
\[
\mathbf{F}'=\left[\begin{array}{ccc}
\mathbf I & | & \mathbf G\end{array}\right]=\left[\begin{array}{cccccccc}
1 &  &  &  & G_{1,1} & \dots & G_{1,N-M}\\
 & 1 &  &  & G_{2,1} & \dots & G_{2,N-M}\\
 &  & \dots &  & \dots & \dots & \dots\\
 &  &  & 1 & G_{M,1} & \dots & G_{M,N-M}
\end{array}\right]
\]
The structure of the linear constraint induced by the row echelon
representation suggests to split the $\bm w$ variable into two sets of
variables: the first $M$ variables (dependent) and a second set of
$N-M$ variables (independent). To do so, we define:
\[
{\bm w}=:({\bm w}^{(d)},\bm{w}^{(i)})=w_{1}^{(d)},\dots,w_{M}^{(d)},w_{1}^{(i)},\dots,w_{N-M}^{(i)}
\]
where ${\bm w}^{(d)}\in\mathbb{R}^{M}$, and ${\bm w}^{(i)}\in\mathbb{R}^{N-M}$.
The linear constraint in equation \eqref{eq:y=Aw} now reads:
\[
{\bm w}^{(d)}+\mathbf G{\bm w}^{(i)}={\bm y}'
\]
where ${\bm y}'\in\mathbb{R}^{M}$ is the transformed measurement
vector. Assuming, as in the previous subsection, a Gaussian prior
for the $\bm{w}^{(i)}$ variables, there follows a Gaussian statistics
on the ${\bm w}^{(d)}$ variables with consistent moments:
\begin{eqnarray*}
{\bm \bar{\bm w}}^{(d)}=-\mathbf G\bar{{\bm w}}^{(i)}+{\bm y'} \\
\bm \Sigma_{{\bm w}^{(i)}}=\left( \mathbf D_{{\bm w}^{(i)}}+\mathbf G^{T}\mathbf D_{{\bm w}^{(d)}}\mathbf G\right)^{-1},\quad\bm \Sigma_{{\bm w}^{(d)}}=\mathbf G\bm\Sigma_{{\bm w}^{(i)}}\mathbf G^{T}
\end{eqnarray*}

where the $\bm D_{{\bm w}^{(d)}}, \bm D_{{\bm w}^{(i)}}$ are the diagonal matrices
whose entries are the inverses of the variances of the Gaussian priors associated with the dependent and independent
variables, respectively. We
note that from the previous relations the moments of the dependent
${\bm w}^{(d)}$ variables can be directly related to those of the
${\bm w}^{(i)}$ variables, which allows us to compute everything
in terms of the inverse of a smaller matrix of size $\left(N-M\right)\times(N-M)$
compared to the finite temperature case.

At this point, the parameters $\bm a$ and $\bm d$ can be updated by moment matching as described in section \ref{sec:ep}.

\section{Moments of the spike-and-slab prior}
\label{app:tilted_mom}
We use the spike-and-slab prior, defined as:
\begin{equation}
\psi(w_n)=(1-\rho)\delta(w_n)+\frac{\rho}{\sqrt{2\pi\lambda}}e^{-\frac{w_n^2}{2\lambda}},
\end{equation}
where $\rho=K/N$ is the density of the signal $\bm w$.
For spike-and-slab priors, the $n$-th marginal of the tilted distribution \eqref{eq:tilted_Q} is given by:
\begin{equation}
Q^{(n)}(w_n)=\frac{1}{Z_{Q^{(n)}}}\tilde{Q}_n(w_n)\psi_n(w_n)
\label{marginalized_tilted}
\end{equation}
in which:
\begin{equation}
\tilde{Q}_n(w_{n})=\frac{1}{\sqrt{2\pi\Sigma'_{n}}}e^{-\frac{(w_{n}-\bar w'_n)^{2}}{2\Sigma'_{n}}},
\end{equation}
where we have denoted $\left(\bar{w}_{(n)}\right)_n$ and $\left(\Sigma_{(n)}\right)_{n,n}$ by $\bar{w}'_n$ and $\Sigma'_n$, respectively, in order to simplify the notation.

The partition function in \eqref{marginalized_tilted} reads:
\begin{equation}
Z_{Q^{(n)}} = (1-\rho)\frac{1}{\sqrt{2\pi\Sigma'_{n}}}e^{-\frac{\bar w_{n}^{'2}}{2\Sigma'_{n}}}+\frac{\rho}{\sqrt{2\pi(\lambda+\Sigma'_{n})}}e^{-\frac{1}{2}\frac{\bar w_{n}^{'2}}{\lambda+\Sigma'_{n}}},
\end{equation}
and the first and second moment of $w_{n}$ with respect to $Q^{(n)}$ are given by:
\begin{equation}
\langle w_{n}\rangle_{Q^{(n)}}=\frac{1}{Z_{Q^{(n)}}}\frac{\rho}{\sqrt{2\pi(\lambda+\Sigma'_{(n)})}}\frac{\lambda\bar w'_{n}}{\lambda+\Sigma'_n} e^{-\frac{1}{2}\frac{\bar w_{n}^{'2}}{\lambda+\Sigma'_{n}}},
\end{equation}
and by:
\begin{equation}
\langle w _{n}^{2}\rangle_{Q^{(n)}}=\frac{1}{Z_{Q^{(n)}}}\frac{\rho}{\sqrt{2\pi(\lambda+\Sigma'_{n})}}\left(\frac{\lambda\Sigma'_{n}(\lambda+\Sigma'_{n})+\lambda^{2}\bar w_{n}^{'2}}{(\lambda+\Sigma'_{n})^{2}}\right)e^{-\frac{1}{2}\frac{\bar w_{n}^{'2}}{\lambda+\Sigma'_{n}}},\quad
\end{equation}
respectively.

\section{Learning of the density parameter of the prior}
\label{app:rho_learn}

\subsection{The EP Free Energy function}
Let us consider equation \eqref{eq:tilted_Q} and rewrite it through equation \eqref{eq:approx_Q} as:
\begin{eqnarray}
Q^{(n)}({\bm w}|\mathbf{F},{\bm y}) & := & \frac{1}{Z_{Q^{(n)}}}e^{-\beta\frac{\left({\bm y}-\mathbf{F}\boldsymbol{w}\right)^{T}\cdot\left({\bm y}-\mathbf{F}\boldsymbol{w}\right)}{2}}\psi_{n}(w_{n})\prod_{l\ne n}\phi_{l}(w_{l};a_{l,}d_{l})\nonumber \\
& = & \frac{Z_Q}{Z_{{Q}^{(n)}}}Q({\bm w}| \mathbf{F},{\bm y})\frac{\psi_{n}(w_{n})}{\phi_n(w_n)}\,.
\end{eqnarray}
We define:
\begin{equation}
  Z_{EP} = Z_Q \prod_{n=1}^N \frac{Z_{Q^{(n)}}}{Z_Q}=\frac{\prod_{n=1}^N Z_{Q^{(n)}}}{Z^{N-1}_Q},
\end{equation}
from which the so-called \emph{EP free energy} follows:
\begin{equation}
  F_{EP} = (N-1) \log Z_Q - \sum_{n=1}^N \log Z_{Q^{(n)}}.
\end{equation}
The converged means $\bm{a}$ and variances $\bm{d}$ of EP, which fulfill the moment matching conditions \eqref{eq:mom_match} for all $n=1,\dots,N$, are fixed points of the EP free energy, where the latter are obtained from:
\begin{eqnarray}
\label{eq:fep_fixed_points_a}
 0 & = & \frac{\partial F_{EP}}{\partial a_n} = (N-1)\langle w_n \rangle_Q-\sum_{l\neq n}\langle w_n \rangle_{Q^{(l)}},\\
 \label{eq:fep_fixed_points_b}
 0 & = & \frac{\partial F_{EP}}{\partial d_n} = (N-1)\langle w_n^2 \rangle_Q-\sum_{l\neq n}\langle w_n^2 \rangle_{Q^{(l)}},
\end{eqnarray}
for $n=1,\dots,N$. In order to show this, we shall prove that the moment matching conditions imply that \eqref{eq:fep_fixed_points_a} and \eqref{eq:fep_fixed_points_b} are satisfied. We have for $\langle w_n \rangle_{Q^{(l)}}$ and $\langle w_n^2 \rangle_{Q^{(l)}}$:
\begin{eqnarray*}
\langle w_n^\alpha \rangle_{Q^{(l)}}&=& \int \frac{Z_Q}{Z_{Q^{(l)}}}Q(\bm w)\frac{\psi_l(w_l)}{\phi_l(w_l)}w_n^\alpha d\bm w = \int  Q(\bm w)\frac{Q^{(l)}(w_l)}{Q(w_l)}w_n^\alpha d \bm w =\\
 & = & \left\langle \int_{-\infty}^{+\infty} \frac{Q(w_n,w_l)}{Q(w_l)} w_n^\alpha dw_n \right\rangle_{Q^{(l)}(w_l)}, \quad \alpha=1,2
\end{eqnarray*}
In the last equality, for $\alpha=1$ ($\alpha=2$), the integral being averaged w.r.t. ${Q^{(l)}(w_l)}$ is the first (second) moment of $w_n$, conditioned on $w_l$ and computed w.r.t. $Q$. These moments depend on $w_l$ through the mean (squared mean) of $Q(w_n|w_l)$. As such mean depends linearly on $w_l$, $\langle w_n \rangle_{Q(w_n|w_l)}$ and $\langle w^2_n \rangle_{Q(w_n|w_l)}$ depend on $w_l$ linearly and quadratically, respectively. Therefore, for $\alpha=1,2$, by the moment matching conditions, we have that:
\begin{eqnarray}
\left\langle \int_{-\infty}^{+\infty} \frac{Q(w_n,w_l)}{Q(w_l)} w_n^\alpha dw_n \right\rangle_{Q^{(l)}(w_l)}=\left\langle \int_{-\infty}^{+\infty} \frac{Q(w_n,w_l)}{Q(w_l)} w_n^\alpha dw_n \right\rangle_{Q(w_l)},
\end{eqnarray}
implying that $\langle w_n^\alpha \rangle_{Q^{(l)}}=\langle w_n^\alpha \rangle_{Q}$ and thus that the conditions \eqref{eq:fep_fixed_points_a} and \eqref{eq:fep_fixed_points_b} are identically fulfilled.

\subsection{Learning of the density}
We are interested in finding the maximum likelihood value of the density parameter $\rho$ which appears in the prior factors $\psi_n(w_n)$.
The likelihood of the parameters of the prior is given by:
\begin{eqnarray}
P(\bm{y}|\rho,\lambda) &=&  \int d\bm{w} P(\bm{y},\bm{w}|\rho,\lambda) = \\
 & = & \int d\bm{w} P(\bm{y}|\bm{w})P(\bm{w}|\rho,\lambda) = Z(\rho,\lambda)
\end{eqnarray}
and maximizing this likelihood corresponds to minimizing the associated free energy $F(\rho,\lambda)=-\log Z(\rho,\lambda)$.

At the fixed point of EP, the free energy is approximated by $F_{EP}$ and the parameters can be learned by gradient descent. In particular, we have for the signal density $\rho$:
\begin{equation}
\rho^{(t+1)}\leftarrow \rho^{(t)} - \eta\frac{\partial F_{EP}}{\partial\rho},
\end{equation}
where $t$ denotes the current iteration and $\eta$ is a learning rate. In the simulations of this paper, we have taken $\eta=5\times 10^{-4}$.

The parameters of the prior enter in the EP free energy through the contributions associated with each of the tilted distributions. Such contributions read:
\begin{equation}
F_{Q^{(n)}}=-\log Z_{Q^{(n)}}=-\log\left(\int \tilde{Q}^{(n)}(\bm{w}|\bm{y})\psi_{n}(w_{n})d\bm{w}\right)
\end{equation}
for
\begin{equation}
\tilde{Q}^{(n)}(\bm{w}|\bm{y}) = e^{-\frac{1}{2}({\bm w}-\bar{{\bm w}}_{(n)})^{T}\Sigma_{(n)}^{-1}({\bm w}-\bar{{\bm w}}_{(n)})} .
\end{equation}
Therefore, we have for $\frac{\partial F_{EP}}{\partial\rho}$:
\begin{equation}
\frac{\partial F_{EP}}{\partial\rho}=\sum_{n=1}^{N}\frac{\partial F_{Q^{(n)}}}{\partial\rho},
\end{equation}
where:
\begin{equation}
\frac{\partial F_{Q^{(n)}}}{\partial\rho}=-\frac{1}{Z_{Q^{(n)}}}\int\tilde{Q}_{n}(w_n)\frac{\partial \psi_{n}}{\partial \rho}(w_n)dw_n,
\end{equation}
and:
\begin{equation}
\frac{\partial \psi_{n}}{\partial \rho}(w_n)=-\delta(w_n)+\frac{1}{\sqrt{2\pi\lambda}}e^{-\frac{w_n^{2}}{2\lambda}},
\end{equation}
yielding:
\begin{equation}
\frac{\partial F_{EP}}{\partial\rho}=\sum_{n=1}^{N}\frac{\frac{1}{\sqrt{2\pi\Sigma_{n,n}}}e^{-\frac{\bar{w}_{n}^{2}}{2\Sigma_{n,n}}}-\frac{1}{\sqrt{2\pi(\lambda+\Sigma_{n,n})}}e^{-\frac{1}{2}\frac{\bar{w}_{n}^{2}}{\lambda+\Sigma_{n,n}}}}{(1-\rho)\frac{1}{\sqrt{2\pi\Sigma_{n,n}}}e^{-\frac{\bar{w}_{n}^{2}}{2\Sigma_{n,n}}}+\frac{\rho}{\sqrt{2\pi(\lambda+\Sigma_{n,n})}}e^{-\frac{1}{2}\frac{\bar{w}_{n}^{2}}{\lambda+\Sigma_{n,n}}}}.
\end{equation}

By taking the derivative w.r.t. $\rho$ one more time, we see that $F_{EP}$ is a strictly convex function of $\rho$ for $\lambda>0$:
\begin{equation}
    \frac{\partial^2 F_{EP}}{\partial\rho^2} = \sum_{n=1}^{N}\left[\frac{\frac{1}{\sqrt{2\pi\Sigma_{n,n}}}e^{-\frac{\bar{w}_{n}^{2}}{2\Sigma_{n,n}}}-\frac{1}{\sqrt{2\pi(\lambda+\Sigma_{n,n})}}e^{-\frac{1}{2}\frac{\bar{w}_{n}^{2}}{\lambda+\Sigma_{n,n}}}}{(1-\rho)\frac{1}{\sqrt{2\pi\Sigma_{n,n}}}e^{-\frac{\bar{w}_{n}^{2}}{2\Sigma_{n,n}}}+\frac{\rho}{\sqrt{2\pi(\lambda+\Sigma_{n,n})}}e^{-\frac{1}{2}\frac{\bar{w}_{n}^{2}}{\lambda+\Sigma_{n,n}}}}\right]^2,
\end{equation}
which guarantees that the sought value of $\rho$ is unique at fixed $\bar{w}_{n}$ and $\Sigma_{n,n}$.

\newcommand{\newblock}{}
\bibliography{./csep.bbl}
\end{document}